\documentclass[sigplan,screen]{acmart}

\usepackage{amsmath}
\usepackage{color}
\usepackage{epsfig}
\usepackage{multirow}
\usepackage{graphicx}
\usepackage{pifont}
\usepackage{hyperref}
\usepackage{color}
\usepackage{booktabs}
\usepackage{mdwlist}
\usepackage{array}
\usepackage{listings}
\usepackage{enumitem}
\usepackage{algorithm}
\usepackage[noend]{algpseudocode}
\usepackage{rotating}

\usepackage[labelfont=bf]{subfig}
\usepackage[font=bf]{caption}

\renewcommand{\algorithmicensure}{\textbf{Output:}}

\newcommand{\pname}[1]{{\textsc{RILOD}}{#1}}

\AtBeginDocument{%
  \providecommand\BibTeX{{%
    \normalfont B\kern-0.5em{\scshape i\kern-0.25em b}\kern-0.8em\TeX}}}

\copyrightyear{2019} 
\acmYear{2019} 
\acmConference[SEC 2019]{The Fourth ACM/IEEE Symposium on Edge Computing}{November 7--9, 2019}{Arlington, VA, USA}
\acmBooktitle{The Fourth ACM/IEEE Symposium on Edge Computing (SEC 2019), November 7--9, 2019, Arlington, VA, USA}
\acmPrice{15.00}
\acmDOI{10.1145/3318216.3363317}
\acmISBN{978-1-4503-6733-2/19/11}

\begin{document}

\title{RILOD: Near Real-Time Incremental Learning for Object Detection at the Edge}

\author{Dawei Li}
\email{dawei.l@samsung.com}
% \author{Serafettin Tasci}
\affiliation{%
  \institution{Samsung Research America}
  \city{Mountain View}
  \state{CA}
}

\author{Serafettin Tasci}
\email{s.tasci@samsung.com}
% \author{Serafettin Tasci}
\affiliation{%
  \institution{Samsung Research America}
  \city{Mountain View}
  \state{CA}
}

\author{Shalini Ghosh}
\email{shalini.ghosh@samsung.com}
% \author{Serafettin Tasci}
\affiliation{%
  \institution{Samsung Research America}
  \city{Mountain View}
  \state{CA}
}

\author{Jingwen Zhu}
\authornote{Work was done when the author was a full-time employee with SRA.}
\email{zhujingwen8942@gmail.com}
% \author{Serafettin Tasci}
\affiliation{%
  \institution{Apple Inc}
  \city{Cupertino}
  \state{CA}
}

\author{Junting Zhang}
\authornote{Work was done when the author was a summer intern with SRA.}
\email{juntingz@usc.edu}
% \author{Serafettin Tasci}
\affiliation{%
  \institution{University of Southern California}
  \city{Los Angeles}
  \state{CA}
}

\author{Larry Heck}
\email{larry.h@samsung.com}
% \author{Serafettin Tasci}
\affiliation{%
  \institution{Samsung Research America}
  \city{Mountain View}
  \state{CA}
}

\renewcommand{\shortauthors}{Li and Tasci, et al.}
% \begin{abstract}
%   A clear and well-documented \LaTeX\ document is presented as an
%   article formatted for publication by ACM in a conference proceedings
%   or journal publication. Based on the ``acmart'' document class, this
%   article presents and explains many of the common variations, as well
%   as many of the formatting elements an author may use in the
%   preparation of the documentation of their work.
% \end{abstract}
\begin{abstract}

Object detection models shipped with camera-equipped edge devices cannot cover the objects of interest for every user. Therefore, the incremental learning capability is a critical feature for a robust and personalized object detection system that many applications would rely on. In this paper, we present an efficient yet practical system, \pname{}, to incrementally train an existing object detection model such that it can detect new object classes without losing its capability to detect old classes. The key component of \pname{} is a novel incremental learning algorithm that trains end-to-end for one-stage deep object detection models only using training data of new object classes. Specifically to avoid catastrophic forgetting, the algorithm distills three types of knowledge from the old model to mimic the old model's behavior on object classification, bounding box regression and feature extraction. In addition, since the training data for the new classes may not be available, a real-time dataset construction pipeline is designed to collect training images on-the-fly and automatically label the images with both category and bounding box annotations. We have implemented \pname{} under both edge-cloud and edge-only setups. Experiment results show that the proposed system can learn to detect a new object class in just a few minutes, including both dataset construction and model training. In comparison, traditional fine-tuning based method may take a few hours for training, and in most cases would also need a tedious and costly manual dataset labeling step.

\end{abstract}





\maketitle

\section{introduction}
\label{sec:intro}

Object detection is the computer vision task that identifies and localizes instances of semantic objects of a certain class (such as humans, buildings, or cars) in images or video frames. On edge devices that are equipped with advanced cameras such as smart-phones, self-driving cars, mobile robots and drones, object detection has been the backbone for many exciting applications including augmented reality, autopilot, mobile shopping, etc. These successful applications were made possible due to recent breakthroughs in deep learning based object detection, especially the one-stage object detection frameworks like YOLO~\cite{redmon2016you}, SSD~\cite{liu2016ssd} and RetinaNet~\cite{lin2018focal}, which enabled on-device inference for applications with real-time requirement. 

In this paper, we focus on a challenging but critical problem for object detection at the edge, i.e., the incremental learning of new object classes. A pre-trained deep learning model shipped with the edge devices could cover up to hundreds of object classes with reasonably good accuracy. However, each end user may have his/her personalized objects of interest that are not included in the pre-trained model, and it would not be feasible to train a general model that covers every user's interest. Meanwhile, new objects of interest may appear from time to time (e.g., the emergence of MP3 Walkman to replace the old tape Walkman), and there's no way to predict those future object classes.

To incrementally add new object classes into the model, a straightforward way is to fine-tune the model with training data from both old and new classes. However, this naive method would not work in practice. First, due to privacy issues, the training data for old classes of the deployed model may not be available with a high possibility. Second, even the old data is available, training using a dataset containing all classes is time consuming as the training efficiency is highly correlated with the amount of training data required. For object detection at the edge, learning to detect new objects incrementally in a timely and quality-conscious manner is crucial in many application scenarios. Users of edge devices may encounter new objects of interest anywhere anytime and delayed learning on those objects would have a negative impact on the user experience. For mobile robots and drones performing military or disaster relief missions, this could play a decisive role on saving lives and properties. 

On the other hand, deep neural networks are known to suffer from ``catastrophic forgetting"~\cite{kirkpatrick2017overcoming} --- this is the phenomenon where, when training using back propagation and the standard cross-entropy loss with only training data of the new classes, the model can forget its learned knowledge on the old classes, e.g., for the object detection model, it can no longer detect objects belonging to the old classes. To deal with the ``catastrophic forgetting" problem, deep learning researchers have proposed effective learning methods such as elastic weight consolidation (EWC) ~\cite{kirkpatrick2017overcoming} and learning without forgetting (LwF) ~\cite{li2017learning}. Those methods have focused on the object classification problem, and none of them can be directly applied to the object detection problem, especially the one-stage object detection models which simultaneously predict the object classes and the bounding box locations. 

We address the ``catastrophic forgetting" problem of incremental learning for one-stage object detection using a novel loss function based on the knowledge distillation technique~\cite{hinton2015distilling}. The fundamental idea is to discourage the change of predictions for the old classes including the model output of both classification and bounding box regression. Specifically, when training only with new-class data, we keep a copy of the old model to generate the output of the new data for both classification (i.e., the probability of belonging to each of the old classes) and bounding box location (i.e., the coordinates of each detected object in the image), and use distillation loss terms to minimize the changes in the new model (see Figure~\ref{fig:method}). By this way, we force the old and new models to reach a consensus on predicting the old classes. Moreover, we add a feature distillation loss term to prevent dramatic changes of features extracted from a middle neural network layer, which further reduces the catastrophic forgetting and improves the model accuracy.

Unlike object classification task, the training data for object detection of a new class may not always be available since it requires both class level labels and accurate bounding boxes, annotating which is a costly and laborious job. In the absence of training data, we need to automatically build a reasonably good training dataset on-the-fly. To this end, we propose a real-time training dataset construction method for the new object classes. We show that the created dataset has accurately annotated bounding boxes around the objects corresponding to the new classes, and thus can be directly used to incrementally train the detection model.

To summarize, we make the following contributions: 
\setlist{nolistsep}
\begin{itemize}[noitemsep]
    \item The first end-to-end incremental learning system for object detection at the edge in near real-time. 
    \item A new algorithm and loss function for incremental learning of one-stage object detection models to solve the ``catastrophic forgetting" problem.
    \item On-the-fly automated dataset construction algorithm to create training data for new classes with both classification and bounding box labels.
    \item Extensive experiments that demonstrate the effectiveness and efficiency of each system component --- we also investigate some trade-offs to optimize the overall system performance.
    \item Prototype implementations of the end-to-end system on both edge-cloud and edge-only architectures. For the edge-only implementation, we develop and deploy the system completely on Jetson TX2 embedded AI platform --- we show that learning of a new class completes in just a few minutes, including both dataset construction and model training.
\end{itemize}

% (\textcolor{red}{Shalini: Another challenge to highlight here is efficiency -- even if we're able to do LwF, the model has to be relatively small and inference fast.})

% (1) System that work for real-time learning of detector
% (2) Learning algorithm
% (3) Dataset construction algorithm
% (4) Build Prototype System and Evaluation

The rest of the paper is organized as follows: We review related work in Section 2. The system overview is described in Section 3. We present our efficient incremental learning method for object detection in Section 4. In Section 5, we introduce the automatic data collection and labeling algorithm. The evaluation and prototype implementation is given in Section 6. Finally we conclude the paper in Section 7.

\section{Related Work}
\label{sec:related}

\begin{figure*}[tph]
%\vspace{-3ex}
  \centering
  \includegraphics[width=2\columnwidth]{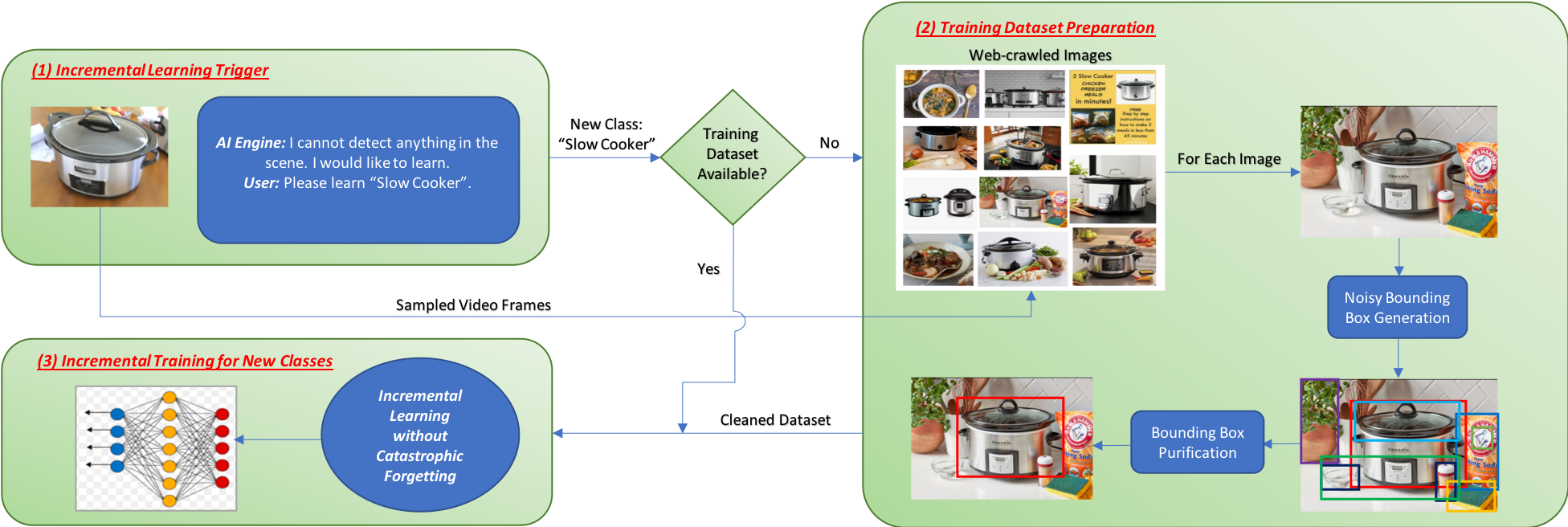}
  %, height=2.8in]{./Figures/Picture_test.png}
  \vspace{2ex}
\caption{System overview of \pname{}. It includes 3 main components: (1) Incremental learning trigger which incorporates the logic on when the incremental learning will start and what classes will be learned (one or more new classes can be learned at a time), (2) Training dataset preparation which downloads, labels and purifies the training images for the new classes in case that there's no available training dataset, and (3) Incremental training for new classes which applies our novel and efficient incremental learning algorithm to incrementally train the model with only the training data of the new classes while maintaining its knowledge on the old classes. Components (2) and (3) may run on a remote cloud server or locally on a mobile GPU for a fully on-device system, and both have been implemented for prototype system evaluation. }
\label{fig:overview}
\end{figure*}

\textbf{Deep Learning for Edge Vision:} Deep learning, especially convolutional neural networks (CNN), has been widely used for a variety of computer vision application on the edge devices due to the recent development of light-weight deep learning SDKs~\cite{tensorFlow, lane:hotmobile2015, CaffeAndroid, DeepBeliefSDK}. Researchers have focused on reducing the computation cost on the edge devices by compressing pre-trained deep learning models~\cite{han_compression, han:mobisys2016, lane:ipsn2016, chen2017learning}, designing lightweight neural network architectures~\cite{squeezenet, howard2017mobilenets}, or hardware accelerations~\cite{huynh2017deepmon, wang2017dlau} --- these papers have not addressed the incremental learning problem in edge vision. 

% None of those works has focused on the problem of learning over the mobile devices Our work focuses on addressing the efficient continual learning problem 

% Recent studies~\cite{han:mobisys2016, han_compression, squeezenet, tucker_cnn} have proposed 
% optimization methods like model compression to trade off the CNN accuracy for 
% resource usage. Our analysis on the compression method 
% demonstrate that these techniques still have significant performance and energy overheads on mobiles. 

% Recent work~\cite{albericio:isca2016, chen:asplos2014, cui:eurosys2016, lane:ipsn2016, li:sc2016, wa:isca2016, liu:isca2016, ovtcharov:whitepaper} 
% proposed using hardware accelerators like GPUs and ASIC %application-specific integrated circuits (ASICs) 
% to accelerate the deep processing. However, these hardware devices 
% are not widely available on commodity mobile devices~\cite{rastegari:arxiv2016}.
% Our work focuses on building energy-efficient deep processing system for 
% commodity mobiles including those low-end mobile phones that may not have advanced
% hardware. And we wish 
% to study the impact of the accelerator hardware on the tradeoff between energy-efficiency
% and data privacy in the future.  

\textbf{Deep Incremental Learning:} To address the ``catastrophic forgetting" problem, deep learning researchers have recently developed methods in several different directions. One important approach is adding regularization~\cite{aljundi2017memory,kirkpatrick2017overcoming} to the neural network weights according to their importance to the old classes, i.e., it discourages the change on important weights using a smaller learning rate. The other major research direction is based the knowledge distillation~\cite{hinton2015distilling}, which uses the new data to distill the knowledge (i.e., the network output) from the old model and mimic its behavior (i.e., generate similar output) when training the new model~\cite{rebuffi2017icarl, li2017learning, shmelkov2017incremental}. Among the related work, only~\cite{shmelkov2017incremental} has focused on the object detection problem. However, their approach works on the Fast-RCNN model~\cite{girshick2015fast}, where bounding boxes proposals are computed using an additional selective search method~\cite{selectiveSearch} --- this approach will not be able to satisfy the real-time inference requirement on edge devices. 

\textbf{Automatic Image Annotation:} A related research topic to our automatic training dataset construction problem is automatic image annotation~\cite{jeon2003automatic, chong2009simultaneous}, which is a process that relies on the computer system assigning metadata (e.g., keywords, caption) to digital images. Image annotation is generally used as a method to provide semantic labels for an image retrieval system. The labeled images is not used as training dataset to train or refine a learning model. Our approach is different from automatic image annotation --- we are trying to automatically build a training dataset on-the-fly by selecting and labeling a subset of training images from noisy web-crawled images. Furthermore, our labeling algorithm not only gives the image-level labels but also object bounding box locations.

\section{System Overview}
\label{sec:overview}

In this section, we give an overview of the proposed \pname{} system which has three major components as illustrated in Figure~\ref{fig:overview}. 

The first component is the incremental learning trigger that runs on the edge devices to listen for the request of learning new classes. There could be different scenarios to initiate a new learning task. In the simplest scenario, the user directly tells the AI engine to learn a new object class that the current AI engine cannot detect. However, this scenario has an unpractical assumption that the user knows what class she/he is interested in and what existing classes the AI engine has learned. 

A more realistic situation is that the user does not know in advance her/his interest or the already learned classes, but would like to learn whenever new classes of interest come across. Therefore, we design an interactive way of initiating a new learning task as shown in Figure~\ref{fig:overview} --- when the camera of an edge device was pointed to a certain object (e.g. a slow cooker), if the AI engine cannot detect the object using the current object-detection model with enough confidence, it prompts the user to notify that it would like to learn the new object category. If the user approves the learning request, she/he could issue a simple \textit{initiate-learning} command and tell the AI engine to start the learning process. For each learning task, one or more new classes can be learned simultaneously.

% Another important problem here is how to designate the name(s) of the new class(es). The straightforward way is that the user would directly tell the AI engine the name of the new object in images captured by the camera (e.g., ``slow cooker"). Alternatively, we could use a reverse image search engine to obtain the object name(s) by using the captured image as a query. 

The second component is a training dataset preparation pipeline which constructs a set of training data in real-time for the new classes to be learned. This component will be triggered if there is not an available training dataset for the new classes to be learned. Specifically, it first downloads relevant noisy images from the Internet using the new class names as query. Then, we purify the downloaded image and generate bounding box annotations for the objects corresponding to the given class names. We will elaborate our solutions in Section~\ref{sec:datalabel}.

The third component is the model training module which learns to detect the new object using either an available training dataset or the dataset prepared by the previous system component. While training with only new-class data reduces the time cost of incremental learning, it increases the risk of $catastrophic forgetting$ on the old classes. Our novel incremental learning algorithm is built on top of the powerful knowledge distillation technique which has been used for incremental learning of object classification models\cite{li2017learning}. However, incremental learning for an object detection model is much more complicated than a classification model, since it requires preservation of not only the semantic classification capability but also its knowledge regarding the bounding box prediction. We will discuss the details in Section~\ref{sec:learning}.

We have implemented the end-to-end system with two different system architectures: edge-cloud and edge-only. For the edge-cloud implementation, the dataset construction and model training components run on a remote cloud-based GPU server --- once the training is completed, the new model is downloaded to the edge devices for future inferences. The edge-only implementation is a fully on-device system without relying on the cloud computing resources. We have evaluated and compared both implementations in our evaluation section.

%distills the knowledge~\cite{hinton2015distilling} of 

%is built on top of the powerful learning-without-forgetting technique which was invented for the incremental learning of object classification models. 

%Our novel life-long learning algorithm guarantees that 1)  the  original  training data  for  old  classes  are  no  longer  accessible  when  learning new classes — this could be due to a variety of reasons,e.g., legacy data may be unrecorded, proprietary, too largeto store, or simply too difficult to use in training the modelfor a new task;  2) the system at any time should providea competitive multi-class classifier for the classes observedso far; 3) the model size should remain approximately thesame after learning new classes.

\section{Incremental learning algorithm}
\label{sec:learning}

\begin{figure}[t]
%\vspace{-3ex}
  \centering
  \includegraphics[width=0.92\columnwidth, height=2.2in]{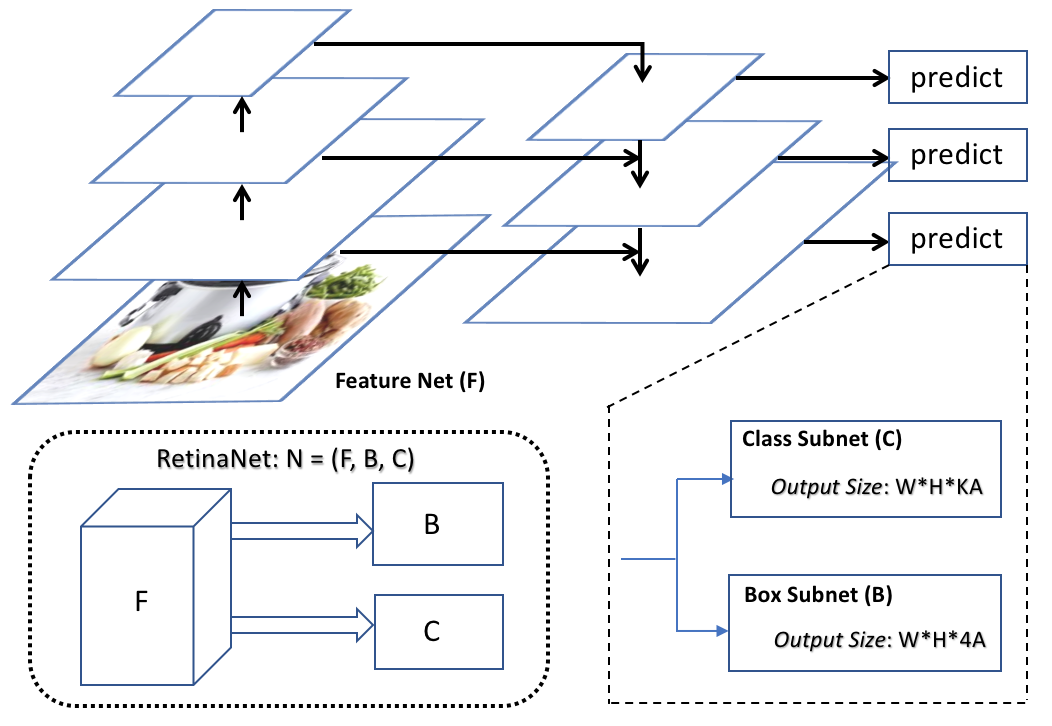}
\caption{A representative one-stage object detection model architecture: RetinaNet~\cite{lin2018focal}. For each cell in a grid of size \textit{W*H}, it predicts (1) the probability that an object exists in each of the \textit{A} anchor boxes from \textit{K} different classes, and (2) the four coordinate offsets for each of the \textit{A} anchor boxes to a ground-truth bounding box (if one exists). }
\label{fig:one-stage}
\end{figure}

\subsection{Recap for One-stage Object Detector}

The goal of object detection is to recognize instances of a predefined set of object classes (e.g. {people, cars, bikes, animals}) and describe the locations of each detected object in the image using a bounding box. Earlier end-to-end deep learning methods adopted a two-stage architecture which first identify a set of possible bounding box locations using a region proposal network and then use a second convolutional neural network to refine the bounding box proposals and classify the object categories~\cite{fasterrcnn}. While two-stage object detectors can achieve very high detection accuracy, they suffer from slow inference speed and thus cannot be deployed for most edge applications which require real-time inference capability. Therefore, recent research has focused on developing one-stage object detection architectures~\cite{lin2018focal, lin2017feature, liu2016ssd, redmon2016you} which run faster yet provide accuracy similar to the two-stage detectors.

The success of one-stage detector is attributed to four critical techniques recently invented by computer vision researchers: grid-based prediction~\cite{redmon2016you}, anchor boxes~\cite{liu2016ssd},  feature pyramids~\cite{lin2017feature} and focal loss~\cite{lin2018focal}. First, the feature maps generated by convolutional neural networks preserve the spatial information of the input image and thus~\cite{redmon2016you} divide the feature maps into grids, and each grid cell is used to directly predict objects (both classification and bounding box) falling into the area. In this manner, there's no need for a box proposal network. Second, regression learning of the four coordinates of a bounding box from random initialization makes the training extremely hard; to address this issue, ~\cite{liu2016ssd} instead predicts bounding box offsets to some pre-defined anchor boxes with varying aspect ratios, which embed some prior information about the shape of candidate objects. Third, to better locate objects of different scales and especially the small ones, ~\cite{lin2017feature} proposed to predict objects with feature maps of different resolutions via a top-down pathway, i.e., the feature pyramids. Finally, the focal loss was proposed~\cite{lin2018focal} to solve the class imbalance problem (e.g., there are too many boxes containing only the image background compared to the boxes actually containing an object). In particular, a scaling factor was added to the cross entropy loss such that the training focuses more on learning hard examples.

In Figure~\ref{fig:one-stage}, we show the basic architecture of the state-of-the-art one-stage object detection method, RetinaNet~\cite{lin2018focal}. RetinaNet is composed of three subnets: a Feature Net (F) for feature map extraction from different resolutions (i.e., different neural network layers), a Class Subnet (C) for object classification, and a Box Subnet (B) for bounding box regression. The Feature Net is also called a backbone network --- it is usually a classification network, pre-trained using a large-scale dataset such as ImageNet~\cite{imagenet_cvpr09}. At inference time, RetinaNet first uses a pre-defined classification probability threshold to decode only boxes with high confidence scores and then uses non-maximum suppression (nms) technique to filter out redundant predictions.

\subsection{Incremental Learning Method}

\begin{figure}[t]
%\vspace{-3ex}
  \centering
  \includegraphics[width=0.92\columnwidth, height=2.2in]{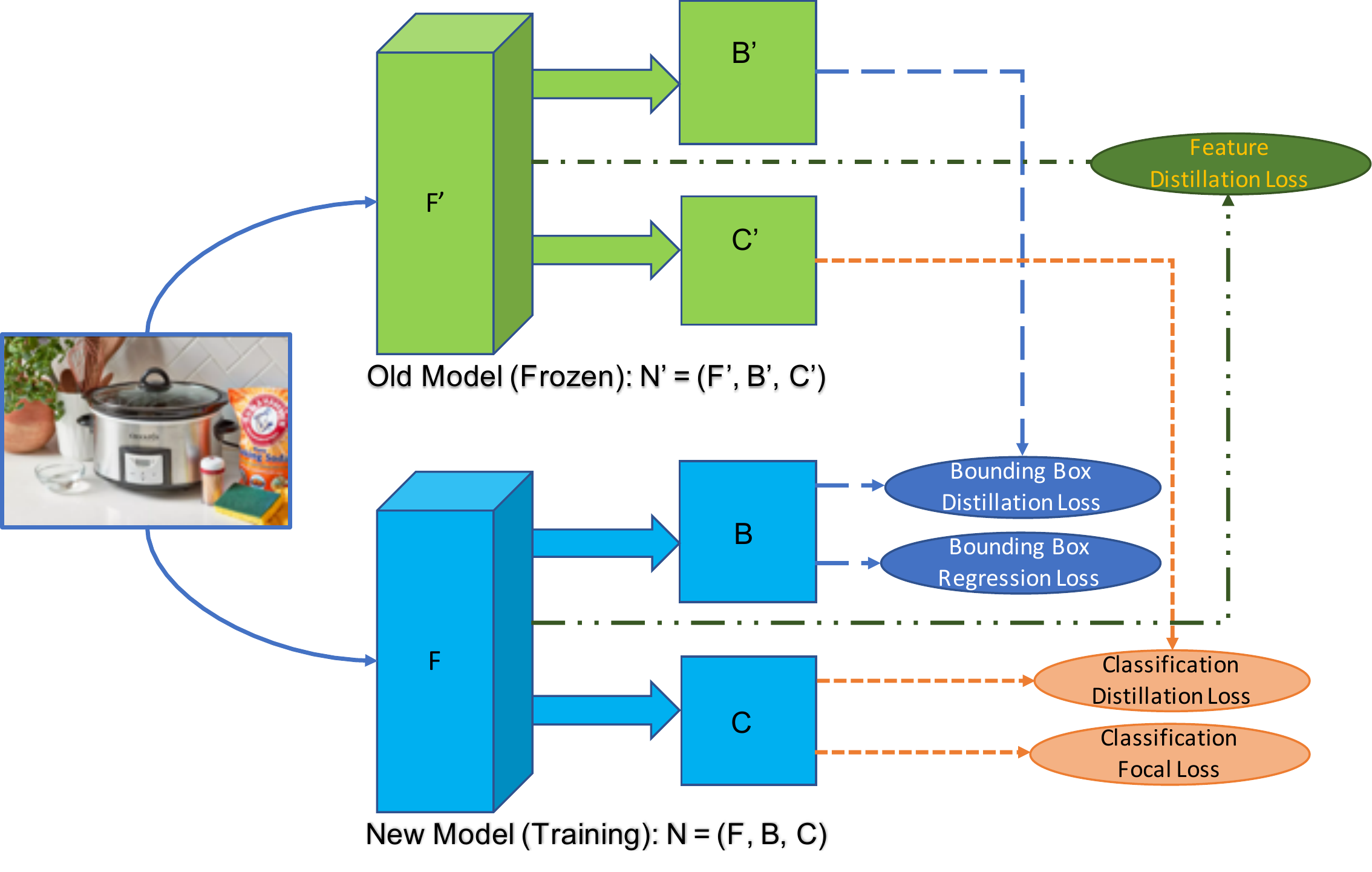}
\caption{The proposed incremental learning method using RetinaNet as an example. }
\label{fig:method}
\end{figure}

The general architecture of our incremental learning algorithm for one-stage deep object detectors is illustrated in Figure~\ref{fig:method}. To detect objects from \textit{n} new classes ($n\ge1$), we first create a copy of the old model as \textit{N'} and then create a new model \textit{N} by expanding the old model's classification subnet to classify \textit{n} more classes by adding \textit{n} neurons to the network's output layer. The weight parameters of the new model are initialized using the corresponding parameters from the old model with the exception of the newly added neurons in the output layer, which are randomly initialized.

To avoid catastrophic forgetting of the old classes while training only with data for new classes, we follow the idea of learning without forgetting (LwF)~\cite{li2017learning} which has been successfully applied to the image classification problem. Specifically, LwF tries to ensure that the classification output for the old classes in the new model (i.e., the vector of the probabilities for the old classes) is close to the output of the old model on the same input image. To achieve this goal, LwF leverages the knowledge distillation~\cite{hinton2015distilling} technique, which uses a modified cross-entropy loss --- instead of using a hard one-hot ground-truth label, it uses the input image's output from the old model as the soft ground-truth label to train the new model. The core idea here is that it discourages the change for the output of old classes. By jointly optimizing the distillation loss on the old classes and the cross-entropy loss on the new classes, LwF achieves good classification performance on predicting both old and new classes.

However, only preserving the classification capability of the old model is not enough for object detection, as the new model would still lose its ability to predict correct bounding boxes for the old classes while training only using the new classes' bounding box labels. To address this issue, we extend the knowledge distillation idea to the object detection task so that it encourages the outputs from both the classification subnet and the box subnet in the new model, in order to approximate the outputs from the old model for the old classes. Furthermore, we argue that applying distillation loss only on the model outputs is not enough to prevent forgetting on old classes. While training only with data for new classes, the intermediate features (i.e., features extracted from a middle layer) which are important for predicting old classes have also been changed during back-propagation. 
%Furthermore, we observed that preserving only the model's outputs does not achieve the desired performance for maintaining the capability on old classes. 
%We argue that this is because while training with only data from the new classes, the intermediate features (i.e., features extracted from a middle layer) which are important for predicting old classes have also been changed with naive back-propagation. 
Therefore, we design a new distillation loss on the intermediate features extracted from the feature net, \textit{F}, so that the features extracted from the new model would not be dramatically different from the old model.

\subsubsection{Detailed Loss Functions}

% (\textcolor{red}{Shalini: One thing to add here may be the axiomatic motivation of the different loss function components, as we had in the slides -- that would motivate why we add each component.})

% Based on the above analysis, the loss function should satisfy the following properties: (1) 

Given the analysis above, the loss function for incremental learning of the object detection model must satisfy the following three properties to avoid \textit{catastrophic forgetting}:

\begin{enumerate}[label=(\roman*)]
\item \textit{Discourage changes on classification output for the old object classes.}
\item \textit{Discourage changes on bounding box locations for predicted objects.}
\item \textit{Prevent dramatic changes for features extracted from intermediate neural network layers.}
\end{enumerate}

% \setlist{nolistsep}
% \begin{itemize}[noitemsep]
%     \item \textbf{Discourage changes on classification output for the old object classes.} 
%     \item \textbf{Discourage changes on bounding box locations for predicted objects.}
%     \item \textbf{Prevent dramatic changes for features extracted from intermediate neural network layers.} 
% \end{itemize}

Given an old object detection model \textit{N'} trained on \textit{m} classes, and a training dataset $D_{new}$ for \textit{n} new classes, our goal is to incrementally train an object detection model \textit{N} which performs object detection on the complete set of $m+n$ classes. The loss function for our incremental detection algorithm is defined as:
\begin{equation}
\begin{aligned}
loss ={} & L_{focal}(Y_n, \hat{Y_n}) \\ 
        & + \lambda_{1}L_{regr}(B_n, \hat{B_n}) \\
        & + \lambda_{2}L_{dist\_clas}(Y_o, \hat{Y_o}) \\
        & + \lambda_{3}L_{dist\_bbox}(B_o, \hat{B_o}) \\
        & + \lambda_{4}L_{dist\_feat}(T, \hat{T})
        % \{F_{x} \in  F_{c} : (|S| > |C|) \\
        % & \cap (\mathrm{minPixels}  < |S| < \mathrm{maxPixels}) \\
        % & \cap (|S_{\mathrm{conected}}| > |S| - \epsilon)\}
\end{aligned}
\end{equation}
where $\lambda_{1}, \lambda_{2}, \lambda_{3}$, and $\lambda_{4} $ are hyper-parameters to balance the weights of different loss terms, and are all set to 1 in our experiments.

The first loss term $L_{focal}$ and the second loss term $L_{regr}$ are the standard loss functions used in~\cite{lin2018focal} to train RetinaNet for the new classes where $Y_n$ represents the ground-truth one-hot classification labels, $\hat{Y_n}$ represents the new model's classification output over \textit{n} new classes, $B_n$ represents the ground-truth bounding box coordinates, and $\hat{B_n}$ represents the predicted bounding box coordinates for the ground-truth objects.
\footnote{The box subnet outputs are actually offsets to the predefined anchor boxes. We simplify the notation in our equations.}

The third loss term $L_{dist\_clas}$ is the distillation loss for the classification subnet similar to that defined in~\cite{li2017learning,shmelkov2017incremental}. Here, $Y_o$ is the output of the frozen old model \textit{F'} for \textit{m} old classes using the new training data, and $\hat{Y_o}$ is the output of the new model \textit{F} for the old classes. Specifically, the loss is calculated by the following equation:
\begin{equation}
L_{dist\_clas}(Y_o, \hat{Y_o}) = \frac{1}{m}\sum_{i=1}^{m} (Y_o^i - \hat{Y_o}^i)^2
\end{equation}

The fourth loss term, $L_{dist\_bbox}$, preserves the old model's capability of correctly predicting bounding boxes for the old classes. Our insight for the bounding box distillation is that when an image is given as input to the old model, even the images does not contain any object belonging to the old classes or not, it would still predict the existence of old-class objects though with relatively low confidence score. We regard the predicted bounding boxes for those relatively low-confidence (but not too low) objects as the old model's bounding box prediction capability. In particular, for each image, we sort the bounding boxes (i.e., the anchor boxes and the coordinate offsets) predicted with the old model based on their classification confidence scores, and select the top \textit{k} bounding boxes $B_o$ as the ground-truth for bounding box distillation loss (see the left sub-figure in Figure~\ref{fig:data-problem} for an example). When incrementally training the new model, we regress the new model's bounding box predictions $\hat{B_o}$ corresponding to the same set of anchor boxes to the ground-truth using smooth L1 loss~\cite{girshick2015fast}.
\begin{equation}
L_{dist\_box}(B_o, \hat{B_o}) = \sum_{j\in\{x,y,w,h\}} smooth_{L_1}(B_o^j - \hat{B_o}^j)
\end{equation}

% (\textcolor{red}{Dawei: need a figure here, train 60-class coco first})

Finally, we append a novel feature distillation loss term $L_{dist\_feat}$ to prevent the extracted intermediate features from drastic change. Similar to the bounding box distillation loss, we calculate the smooth L1 loss between the features extracted from the feature net of the old model $T$ and the new model $\hat{T}$.
\begin{equation}
L_{dist\_feat}(T, \hat{T}) = \sum_{l} smooth_{L_1}(T^l - \hat{T}^l)
\label{eq:feat-dist}
\end{equation}

% \subsubsection{Implementation Details}

%  (as shown in Figure~\ref{fig:method})

% To avoid catastrophic forgetting of the old classes while training with only data from new classes, we following the idea of learning without forgetting (LwF)~\cite{li2017learning} which has been successfully applied to the image classificaiton problem. 

\subsubsection{Using Exemplars of Old Classes}
\label{sec:exemplar}

%(\textcolor{red}{Shalini: Motivate the use of exemplars -- why is it ok to store exemplars instead of whole data, how can we smartly select exemplars vs random images, etc.})

%Talk about drawbacks: increase in training time, storage cost, advantages: higher accuracy and less forgetting on old classes

%Methodology: 
%1) similar to iCarl, use images in the center of the class
%2) cluster images and take representatives from each class
%3) randomly select images

The basic \pname{} system assumes that no data from old classes are saved and learning happens only using the new class data, which are readily available or collected using the pipeline described in Section~\ref{sec:datalabel}. This assumption reduces both the expected training time and the storage cost for incremental learning. Moreover, it increases the applicability of our system for cases where old data is not available.

However, in some application scenarios some or all of the data for old classes might be available. In such cases, we can augment a small number of exemplars from this old data to our training set to ensure that at least some information about old classes is incorporated into the learning process. In this way, we can further reduce the forgetting on old classes and achieve a better detection accuracy overall. Note that, even when available, using all old data is not reasonable since it will increase the training latency of the system tremendously and make it unsuitable for real-time incremental object detection. Moreover, in our experiments we observed that using only a few exemplar images per class provides similar accuracy to using all data for old classes (see Section \ref{sec:eval}).  

Exemplar image selection can be performed in various ways: 1) randomly select a fixed number of images from each class, 2) select images such that the average feature vector of exemplars will be closest to the class mean as in \cite{rebuffi2017icarl}, 3) perform clustering for each class and pick a random image from each cluster. In this work, we preferred the last method for exemplar selection to consider in-class variation (e.g. select example images from each dog breed for dog class). In addition, even though we select a fixed number of exemplars from each class, it is also possible to fix the total number of exemplars to prevent the linear increase in exemplar dataset size as new classes are added incrementally.

\section{Data Collection and Labeling}
\label{sec:datalabel}

\begin{figure}[t]
%\vspace{-3ex}
  \centering
  \includegraphics[width=0.92\columnwidth, height=1.4in]{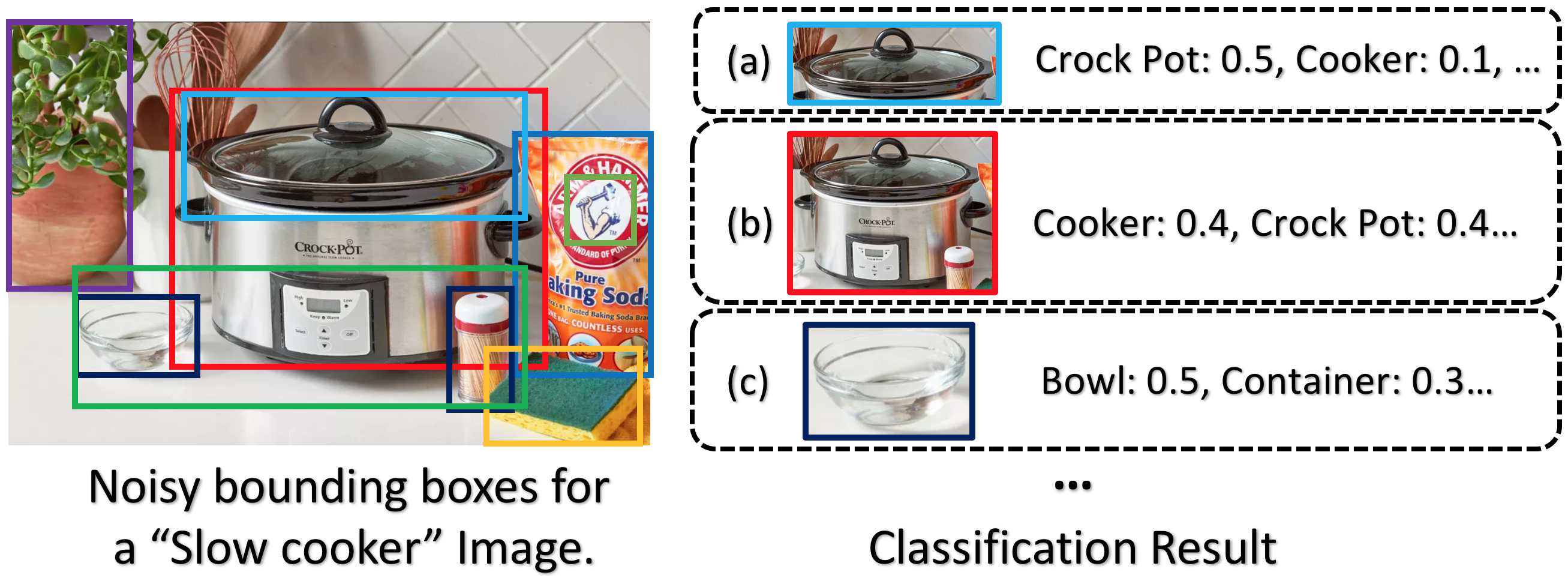}
\caption{An example demonstrating the bounding box labeling problem. First, the labels in the classification model may not match the given new class name. Second, the noisy bounding boxes may overlap with each other (e.g., (a) and (b)) and the ideal box (b) may be eliminated by simply setting a low IoU threshold for NMS. }
\label{fig:data-problem}
\end{figure}

As discussed in Section~\ref{sec:overview}, it is quite likely that there's no existing labeled image for the new object class (e.g., slow cooker), but it is convenient to download a large quantity of relevant but noisy images using search engines (e.g., Google Image Search). We have designed a system module and algorithm to purify the noisy images and annotate the bounding boxes. 

Given a set of downloaded images matching a new class name (e.g., slow cooker), our goal is to automatically and efficiently label the images that do contain the desired objects with bounding boxes. Generally speaking, our system first generates a set of noisy bounding boxes from each downloaded candidate image which can be done using either a class-agnostic method~\cite{selectiveSearch,zitnick2014edge} or an existing deep object detector with a low confidence threshold, then we use a large-scale image classifier to verify each bounding box to filter out the bad box proposals. 

% Even though images with bounding box labels are scarce, there is an abundance of images with class labels such as ImageNet which has over 10 million labeled images for over 10k classes. With a state-of-the-art convolutional architecture (e.g., ResNet~\cite{resnet} or DenseNet~\cite{huang2017densely}), a highly-accurate and efficient classification model can be trained to perform classification over a huge number of classes. Therefore, we argue that once we generate a set of noisy bounding boxes from a candidate image (this can be done using either a class-agnostic method~\cite{selectiveSearch,zitnick2014edge} or an existing deep object detector with a low confidence threshold), then we could use a large-scale classifier to verify each bounding box to filter out the bad box proposals. 

\begin{algorithm}[t]%[htbp]
	\caption{Automatic Dataset Construction }
	\label{alg:data}
	\begin{algorithmic}[1]
		\renewcommand{\algorithmicensure}{\textbf{Input:}}
		\Require the given new class name $l_g$
		\Require the downloaded images $D$ using $l_g$ as query
		\Require the large-scale classification model $M_{cls}$
		\Require the Word2vec model $M_{w2v}$

        \Statex
		\Statex \textit{Voting for ``credible labels" set $S_{cl}$:}
		\State initialize a counter $Ct$ for each label in $M_{cls}$
		\For {each image in $D$}
		\State produce a set of noisy bounding boxes $B$
		\For {each $b$ in $B$ that $size(b)> thr_b$}
		    \State predict top $k$ labels $L_k$ using $M_{cls}$
		    \For{each label $l$ in $L_k$}
		    \State $Ct[l]+=1$
		    \EndFor
		\EndFor
		\EndFor
		\State sort $Ct_l$ and append the top1 label $l_1$ to $S_{cl}$
		\For{each $l_i$ in $sorted(Ct_l)$}
		\If {$cos\_sim(l_i, l_1)+cos\_sim(l_i, l_g)>thr_d$}
		\State $S_{cl}.append(l_i)$
		\EndIf
		\EndFor
		\Return $S_{cl}$
		\Statex
		
		\Statex Purification for final dataset $D_f$
		\For {each image $I$ in $D$}
		
		\For {Each bounding box $b_i$ in $B$}
		\If {Any label in predicted top \textit{k} $\in S_{cl}$}
		\State calculate $ACCS_i$
		\Else
		\State remove $b_i$ from $B$
		\EndIf
		\EndFor
		
		    \If{$size(B)>1$}
    \For{each box pair $b_i, b_j$}
    \If {$overlap(b_i, b_j)>thr_o$}
    % \If{$abs(ACCS_i, ACCS_j)>thr_{accs}$}
    \State remove the box with lower $ACCS$
    % \Else
    % \State remove either one or skip the image
    % \EndIf
    \EndIf
    \EndFor
    \EndIf
    
        \If{$size(B)\ge1$}
    \State add $I$ with $B$ to $D_f$
    \EndIf
		\EndFor
	\Return $D_f$
	\end{algorithmic}
\end{algorithm}

    % \If{$size(B)>1$}
    % \For{each box pair $b_i, b_j$}
    % \If {$overlap(b_i, b_j)>thr_o$}
    % % \If{$abs(ACCS_i, ACCS_j)>thr_{accs}$}
    % \State remove the box with lower $ACCS$
    % % \Else
    % % \State remove either one or skip the image
    % % \EndIf
    % \EndIf
    % \EndFor
    % \EndIf
    
    % \If{$size(B)\ge1$}
    % \State add $I$ with $B$ to $D_f$
    % \EndIf

However, there are two vital challenges that must be addressed as shown in Figure~\ref{fig:data-problem}. The first one is label mismatch problem in which the provided new class name does not have an exact match with the corresponding label in the large-scale classification model, e.g., a ``slow cooker" may be labeled as ``crock pot". In addition, the classification model is not perfect and it may classify a ``slow cooker" as ``pressure cooker" or just its super-class ``cooker". To solve these problems, we propose a hybrid voting+semantic method based on two observations: (1) most of the downloaded images contain the objects matching the given class name, and (2) even though not perfect, the large-scale classification model still has a high recognition accuracy, e.g., we use a 11k-class model with 71.2\% top-5 accuracy in our prototype system. Based on these observations, we could safely conclude that a considerable proportion of the images would have one or more bounding boxes predicted with the ``true label" in the classification model. Because of this, the ``true label" can be identified using a voting mechanism that sorts the labels by the number of occurrences in the classification results over the noisy bounding boxes and return the most frequent label as the ``true label". In addition, other classification labels that are semantically close to both the ``true label" and the given class name presumably also refer to the same object category. The semantic similarity can be measured by the distance of two labels' word embeddings such as Word2vec~\cite{mikolov2013distributed}. The ``true label" and the additional semantically verified labels form the set of ``credible labels".

The second challenge is the overlapping problem among bounding boxes with correct labels. Simply setting a low IoU (intersection over union) threshold for NMS can eliminate too many true positive boxes as shown in Figure~\ref{fig:data-problem}. To deal with the overlapping-box problem, we introduce the \textit{accumulated classification confidence score} (ACCS) to determine which bounding box should be retained from two overlapped boxes. Specifically, the ACCS for a bounding box is defined as the sum of the predicted classification confidence scores for labels belonging to the generated set of ``credible labels". If the the overlapping between two boxes is larger than a threshold,  we can discard the one with lower ACCS.
% Otherwise, the two boxes are indistinguishable by the classifier and we may randomly pick one box or just discard the image since we don't have enough confidence to make the correct decision. 
Additionally, we ignore too small bounding boxes. The reason is that when you use the class name as a query to download images, the object corresponding to the class name is relatively large and clearly visible. Discarding these small boxes before sending them to the classification model also greatly reduces the computational overhead.

The pseudo-code for our dataset construction pipeline is summarized in Algorithm \ref{alg:data}. In out implementation, we set the box size threshold $thr_b$ as 1\% of the image size. The word embedding similarity threshold $thr_d$ is set to 10. The overlapping threshold $thr_o$ of two boxes is set to half the size of the smaller box. $k$ is set to 5.

\textbf{Discussion on Dataset Construction}
It might seem that instead of incrementally training an object classifier using the automatically constructed dataset, the two-stage dataset construction algorithm can be directly employed to detect the new object classes without any training. However, this approach would have more storage and memory cost since it requires the deployment of two deep learning models; a detection model to create the set of noisy bounding boxes and a large-scale classification model for verification of the labels for each bounding box. In addition, it would take significantly more time for inference of a single image compared to single stage object detectors due to the significant time required for classification of box proposals. 
% Therefore, compared to the continual learning method, both storage and computing cost are much higher.

% alg word_dic: count the number of occurrence for each label (in top 5 prediction)
% label_conf_list: the confidence for each label (top 5 labels) predicted of a box
% boxlist: all bounding boxes for an image with their coordinates and label_conf_list
%            [j, label_conf_list, [x1, y1, x2, y2]]
% boxlist_dic: key: img_id, val: boxlist
% true_labels: the label kept
% for each image:
%    get the boxlist
%    check if any of top5 label belong to true_labels, sum the confidence over those labels
%    if there is only one box:
%        return the box
%    else:
%        removeBad
%    
% 		def removeBad(boxes):
% 			removedList = []
% 			mids = copy.deepcopy(boxes)
% 			# print mids, len(ids)
% 			for m in range(len(boxes)):
% 				for n in range(m + 1, len(boxes)):
% 					boxi = boxes[m][-1]
% 					boxj = boxes[n][-1]
% 					areai = (boxi[2] - boxi[0]) * (boxi[3] - boxi[1])
% 					areaj = (boxj[2] - boxj[0]) * (boxj[3] - boxj[1])
% 					coarea = computeArea(boxi[0], boxi[1], boxi[2], boxi[3], boxj[0], boxj[1], boxj[2], boxj[3])
% 					if float(coarea) / min(areai, areaj) > 0.4:
% 						if abs(boxes[m][2] - boxes[n][2]) <= 0.1:
% 							return []
% 						else:
% 							scorem = boxes[m][2]
% 							scoren = boxes[n][2]
% 							if scorem > scoren and boxes[n] in mids:
% 								mids.remove(boxes[n])
% 							if scoren >= scorem and boxes[m] in mids:
% 								mids.remove(boxes[m])
% 			return mids

% We propose 
\section{Evaluation}
\label{sec:eval}

The evaluation of~\pname{} focuses on answering the following questions: (1) How effective and efficient is the proposed incremental learning algorithm? (2) How effective and efficient is the automatic data construction algorithm? (3) How does this complete system work in practice? What is the running time? What are the bottlenecks in overall system design?

\subsection{Evaluation Dataset}

The following two datasets are used to evaluate the object detection accuracy for the incremental learning algorithm.

\begin{table*}[htbp]
	\centering
	\caption{Per-class accuracy of Pascal VOC (\%) for $19+1$ scenario.}
\label{tab:pascal_19}%
	\setlength\tabcolsep{3pt}
\resizebox{\linewidth}{!}{
	\begin{tabular}{|c|c|c|c|c|c|c|c|c|c|c|c|c|c|c|c|c|c|c|c|c|c|}
		\hline
		Method & \begin{sideways}aero\end{sideways} & \begin{sideways}bike\end{sideways} & \begin{sideways}bird\end{sideways} & \begin{sideways}boat\end{sideways} & \begin{sideways}bottle\end{sideways} & \begin{sideways}bus\end{sideways} & \begin{sideways}car\end{sideways} & \begin{sideways}cat\end{sideways} & \begin{sideways}chair\end{sideways} & \begin{sideways}cow\end{sideways} & \begin{sideways}table\end{sideways} & \begin{sideways}dog\end{sideways} & \begin{sideways}horse\end{sideways} &  \begin{sideways}mbike\end{sideways} & \begin{sideways}person\end{sideways} & \begin{sideways}plant\end{sideways} & \begin{sideways}sheep\end{sideways} &
	\begin{sideways}sofa\end{sideways} & \begin{sideways}train\end{sideways} &  
	\begin{sideways}tv\end{sideways} & mAP \\
	\hline
Class 1-19 & 70.6 & 79.4 & 76.6 & 55.6 & 61.7 & 78.3 & 85.2 & 80.3 & 50.6 & 76.1 & 62.8 & 78.0 & 78.0 & 74.9 & 77.4 & 44.3 & 69.1 & 70.5 & 75.6 & - & - \\

All Data & 77.8 & 85.0 & 82.9 & 62.1 & 64.4 & 74.7 & 86.9 & 87.0 & 56.0 & 76.5 & 71.2 & 79.2 & 79.1 & 76.2 & 83.8 & 53.9 & 73.2 & 67.4 & 77.7 & 78.7 & 74.7\\

Catastrophic Forgetting & 0 & 0 & 0 & 0 & 0 & 0 & 0 & 0 & 0 & 0 & 0 & 0 & 0 & 0 & 0 & 0 & 0 & 0 & 0 & 68.9 & 3.4 \\

\hline

w/o Feat-Distill Loss & 61.9 & 78.5 & 62.5 & 39.2 & 60.9 & 53.2 & 79.3 & 84.5 & 52.3 & 52.6 & 62.8 & 71.5 & 51.8 & 61.5 & 76.8 & 43.8 & 43.8 & 69.7 & 52.9 & 44.6 & 60.2\\

w Feat-Distill Loss & 69.7 & 78.3 & 70.2 & 46.4 & 59.5 & 69.3 & 79.7 & 79.9 & 52.7 & 69.8 & 57.4 & 75.8 & 69.1 & 69.8 & 76.4 & 43.2 & 68.5 & 70.9 & 53.7 & 40.4 & \textbf{65.0} \\
	\hline
\end{tabular}
}%
\end{table*}%

\begin{table*}[htbp]
	\centering
	\caption{Per-class accuracy of Pascal VOC (\%) for $10+10$ scenario.}
\label{tab:pascal_10}%
	\setlength\tabcolsep{3pt}
\resizebox{\linewidth}{!}{
	\begin{tabular}{|c|c|c|c|c|c|c|c|c|c|c|c|c|c|c|c|c|c|c|c|c|c|}
		\hline
	Method & \begin{sideways}aero\end{sideways} & \begin{sideways}bike\end{sideways} & \begin{sideways}bird\end{sideways} & \begin{sideways}boat\end{sideways} & \begin{sideways}bottle\end{sideways} & \begin{sideways}bus\end{sideways} & \begin{sideways}car\end{sideways} & \begin{sideways}cat\end{sideways} & \begin{sideways}chair\end{sideways} & \begin{sideways}cow\end{sideways} & \begin{sideways}table\end{sideways} & \begin{sideways}dog\end{sideways} & \begin{sideways}horse\end{sideways} &  \begin{sideways}mbike\end{sideways} & \begin{sideways}person\end{sideways} & \begin{sideways}plant\end{sideways} & \begin{sideways}sheep\end{sideways} &
	\begin{sideways}sofa\end{sideways} & \begin{sideways}train\end{sideways} &  
	\begin{sideways}tv\end{sideways} & mAP \\
	\hline
	Class 1-10 & 76.8 & 78.1 & 74.3 & 58.9 & 58.7 & 68.6 & 84.5 & 81.1 & 52.3 & 61.4 & - & - & - & - & - & - & - & - & - & - & -\\

All Data & 77.8 & 85.0 & 82.9 & 62.1 & 64.4 & 74.7 & 86.9 & 87.0 & 56.0 & 76.5 & 71.2 & 79.2 & 79.1 & 76.2 & 83.8 & 53.9 & 73.2 & 67.4 & 77.7 & 78.7 & 74.7\\

Catastrophic Forgetting & 0 & 0 & 0 & 0 & 0 & 0 & 0 & 0 & 0 & 0 & 66.3 & 71.5 & 75.2 & 67.7 & 76.4 & 38.6 & 66.6 & 66.6 & 71.1 & 74.5 & 33.7\\

\hline
w/o Feat-Distill Loss & 67.1 & 64.1 & 45.7 & 40.9 & 52.2 & 66.5 & 83.4 & 75.3 & 46.4 & 59.4 & 64.1 & 74.8 & 77.1 & 67.1 & 63.3 & 32.7 & 61.3 & 56.8 & 73.7 & 67.3 & 62.0 \\

w/ Feat-Distill Loss & 71.7 & 81.7 & 66.9 & 49.6 & 58.0 & 65.9 & 84.7 & 76.8 & 50.1 & 69.4 & 67.0 & 72.8 & 77.3 & 73.8 & 74.9 & 39.9 & 68.5 & 61.5 & 75.5 & 72.4 & \textbf{67.9} \\
	\hline
\end{tabular}
}%
\end{table*}%

\setlist{nolistsep}
\begin{itemize}[noitemsep]
    \item \textit{\textbf{Pascal VOC 2007}}~\cite{pascal-voc-2007}: This is a benchmark dataset for object detection which includes 20 object classes. In total it has 9,963 images collected from Flickr\footnote{\url{https://www.flickr.com/}} photo-sharing web-site and contains 24,640 annotated objects. We use the 5K images in the \textit{train} and \textit{val} splits for training, and the images in the \textit{test} split for validation.
    \item \textit{\textbf{iKitchen}}: We collected and annotated 7K images (6K for training and 1K for validation) belonging to 10 classes of kitchen objects for an internal demonstration. Most of the images are collected using
    % \textcolor{red}{(mobile phones!!! and relatively simple)} 
    smartphones in lab and real-world environment
    to evaluate our system in a more realistic setting~\cite{ng2017machine}. This dataset was also used in the prototype system evaluation.
\end{itemize}

To evaluate the automatic dataset construction algorithm, we simulate the behavior of downloading web images using a class name as query. Specifically:

\setlist{nolistsep}
\begin{itemize}[noitemsep]
    \item regarding \textit{\textbf{Pascal VOC}}, for each of the 20 classes, we use the class name (e.g., chair) as the search keyword of ImageNet\footnote{\url{http://www.image-net.org}} to find the matched synset, and we download 200 images from the synset along with their bounding box annotations. 
    %Similar to web searched images, ImageNet images generally display the synset object as the main object.
    \item regarding \textit{\textbf{iKitchen}}, we use the Google image search to download images in real-time to demonstrate the effectiveness and efficiency of the algorithm in the prototype system evaluation.
\end{itemize}

For all the model accuracy evaluations, we use the standard
mean average precision (mAP) at the threshold of 0.5 IoU as
the evaluation metric~\cite{everingham2010pascal}.

\subsection{Incremental Learning}

The input image for the deep neural network is resized such that the longer side is 1024 and the shorter side is then adjusted proportionally. The initial learning rate for all experiments is set to $1\times10^{-3}$ and we use the Adam method~\cite{kingma2015adam} for optimization. The training is performed on a single NVIDIA Tesla M40 GPU, with batch size of 8. We run all our experiments using PyTorch~\cite{paszke2017pytorch}.

%Random horizontal flipping is used for data augmentation. Standard non-maximum suppression (NMS) with threshold 0.5 is applied for post-processing at test time to remove the duplicate predictions. For each image, we select 64 anchor boxes for \pname training. Empirically we found selecting more anchor boxes (128, 256 etc.) did not provide further performance gain. The $\lambda$ is set to 1.0 for all experiments.

\subsubsection{Experiment Setup For Pascal Dataset}

The model architecture we use for Pascal dataset is RetinaNet with ResNet-50 backbone, and we train each model for 100 epochs. We performed experiments for the following two scenarios: 

\setlist{nolistsep}
\begin{itemize}[noitemsep]
    \item \textbf{19+1:} we train a base model with the first 19 classes and incrementally train the 20th class (i.e., the tv/monitor class). 
    %We run a series of 20 experiments by setting each of the 20 classes as the new class to learn.
    \item \textbf{10+10:} we train a base 10-class model and then learn 10 new classes simultaneously. 
\end{itemize}

The following 4 different learning schemes have been compared:

\setlist{nolistsep}
\begin{itemize}[noitemsep]
    \item \textbf{All Data:} We fine-tune the old model using data from both old and new classes and use the original loss function in RetinaNet without distillation loss terms. 
    %We run a series of 20 experiments by setting each of the 20 classes as the new class to learn.
    \item \textbf{Catastrophic Forgetting:} We train the model using the training data of only new classes and use the original loss function in RetinaNet without distillation loss terms. 
    \item \textbf{w/o Feat-Distill Loss:} We train the model with training data of only new classes and use the proposed loss function except the feature distillation loss term defined in Equation~\ref{eq:feat-dist}.
    
    \item \textbf{w Feat-Distill Loss:} We train the model using training data of only new classes and we use the proposed loss function including the feature distillation loss term. 
\end{itemize}

\subsubsection{Experiment Result For Pascal Dataset}

% \begin{table}[t]
% \caption{Accuracy of Pascal 19+1 scenarios. ($Base_O$: the average accuracy of the base 19 classes with old model; $Base_N$: the average accuracy of the base 19 classes with new model; $New$: the accuracy of the new added class; $Avg$: the average 20-class accuracy of the new model.)}
% \centering
% \footnotesize
% \begin{tabular}{m{1.15cm}|m{1.2cm}|m{1.2cm}|m{1.2cm}|m{1.2cm}}
% % \hline
% % \multicolumn{2}{c|}{\textbf{Mobile Client} (Samsung Galaxy S5)} & \multicolumn{2}{c}{\textbf{Cloud Server} (GPU server)} \\
% \hline
% $+1Class$ & $Base_O$  & $Base_N$  & $New$ &  $Avg$ \\
% \hline
% airplane & \% & \% & \% & \% \\
% bike & \% &  &  &  \\
% bird & \% &  &  &  \\
% boat & \% &  &  &  \\
% bottle & \% &  &  &  \\
% bus & \% &  &  &    \\
% car & \% &  &  &  \\
% cat & \% &  &  &  \\
% chair & \% &  &  &  \\
% cow & \% &  &  &  \\
% dtable & \% &  &  &  \\
% dog & \% &  &  &  \\
% horse & \% &  &  &  \\
% mcycle & \% &  &  &  \\
% person & \% &  &  &  \\
% plant & \% &  &  &  \\
% sheep & \% &  &  &  \\
% sofa & \% &  &  &  \\
% train & \% &  &  &  \\
% tv & \% &  &  &  \\
% \hline 
% \end{tabular}
% \label{tab:19+1_all}
% \end{table}

\begin{figure*}[tph]
%\vspace{-3ex}
  \centering
  \includegraphics[width=2\columnwidth, height=1.8in]{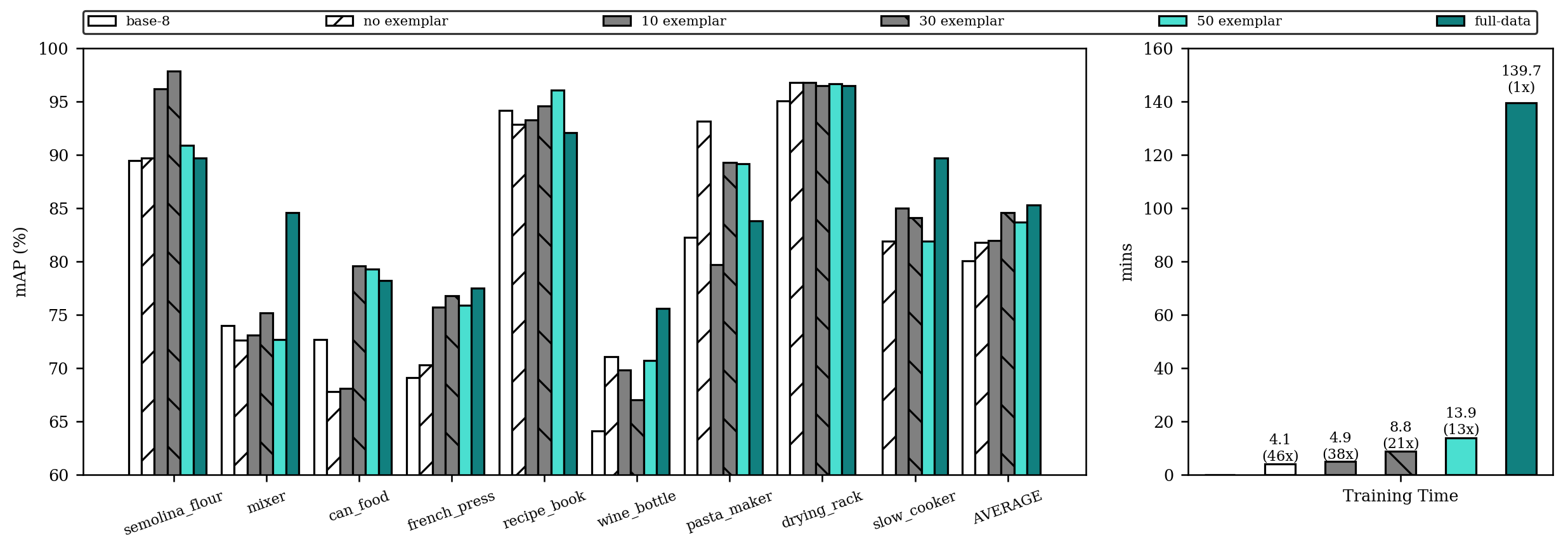}
\caption{Result of adding ``slow cooker" to the base 8 class model on iKitchen Dataset.}
\label{fig:ikitchen_slow_cooker}
\end{figure*}

We present the result of 19+1 scenario in Table~\ref{tab:pascal_19}, and the result of 10+10 scenario in Table~\ref{tab:pascal_10}. In both scenarios, we observe that without the distillation loss terms, the ``catastrophic forgetting" problem occurs and the mAP values for all old classes drops to 0. On the other hand, with the novel loss function we have proposed in this paper, the accuracy on the old classes is preserved while incrementally learning with training images of only the new classes. Even compared to using data of all classes, the average mAP over all classes is only reduced by less than 10\%. In addition, by adding the feature distillation loss term, we observe 4.8\% accuracy increase for the 19+1 scenario, and 5.9\% accuracy increase for the 10+10 scenario. The results demonstrate that the proposed learning algorithm can solve the problem of ``catastrophic forgetting" pretty well even in scenarios when multiple new classes are learned simultaneously.

In the 19+1 scenario, the number of training images for the new ``tv/monitor" class is just 279 which is only 5.5\% of the 5K training dataset for all 20 classes. Even with such a limited amount of data, our model can learn the new class without forgetting the old classes. Moreover, due to its capability of learning with a small number of images, our method can significantly reduce the training time and thus is more suitable for incremental learning in edge applications.

% \begin{figure}[t]
% %\vspace{-3ex}
%   \centering
%   \includegraphics[width=0.92\columnwidth, height=2.2in]{./Figures/m_10.png}
% \caption{\textcolor{red}{Replaced with a figure showing the convergence time mAP of 19+1: continual vs all-data} }
% \label{fig:pascal_time}
% \end{figure}

\subsubsection{Experiment Setup For iKitchen Dataset}

The model architecture we use for iKitchen dataset is RetinaNet with ResNet-18 backbone. We train a base 8-class object detection model by excluding two classes \textit{slow cooker} (SC) and \textit{cocktail shaker} (CS). The mAP for the base 8-class model is 80.1\%.

Then we run the following experiments: (1) 8+SC, (2) 8+CS, (3) 8+SC+CS, and (4) 8+CS+SC. In 3 and 4, we apply the proposed incremental learning method incrementally as each new class is added. Moreover, we study the trade-off between accuracy and training time of adding exemplar data from the old classes as discussed in Section~\ref{sec:exemplar}. For all experiments, we train the model for 10 epochs using the proposed incremental learning algorithm including the feature distillation loss.

% We also compare study the tradeoff of adding exemplar data.

\subsubsection{Experiment Result For iKitchen Dataset}

\begin{figure}[t]
  \centering
  \includegraphics[width=0.92\columnwidth, height=2.2in]{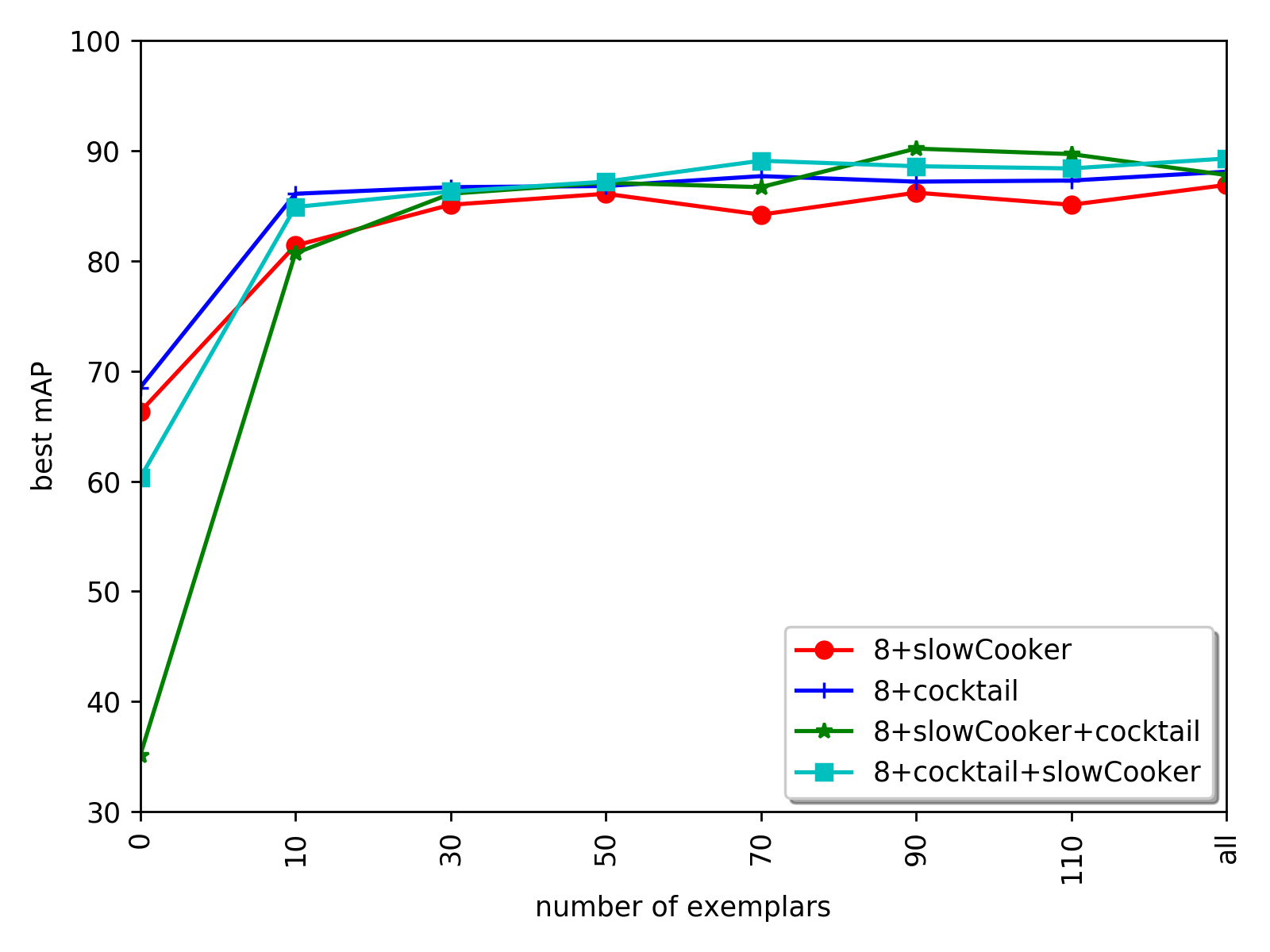}
\caption{The average mAP over all classes with different number of exemplars per class. }
\label{fig:accuracy_exemplar}
\end{figure}

As seen in in Figure~\ref{fig:accuracy_exemplar}, adding exemplars from the old classes can boost the incremental learning accuracy, especially when we are learning new classes sequentially (i.e., the 8+SC+CS and 8+CS+SC scenarios). Even by adding just 10 exemplars per old class, the mAP increases 15\% to 40\%. As more exemplars are added, we don't see further significant accuracy increase. Using all the training data from old classes can be seen as a special case of adding exemplars. 

In Figure~\ref{fig:ikitchen_slow_cooker}, we show the detailed accuracy for each object class in the 8+SC scenario for different number of exemplars per class (left sub-figure). First of all, we see that all the old classes maintain a good accuracy after learning the new class. In addition, we see that the accuracy for some of the base 8 classes also increases after adding the new class. The reason could be that as the model sees more diverse training samples, it can learn better features that provides better boundaries for discriminating different object classes. We have also measured the training time and the speed-up of using different number of exemplars (right sub-figure). Compared with the baseline of learning with all training data, we can achieve 38x speed-up with only 4\% accuracy loss (see the 10 exemplar case).

\begin{table*}[t]
\caption{Credible Labels for PASCAL Dataset.}
\centering
\footnotesize
\begin{tabular}{m{2.5cm}|m{14cm}}
\hline
 \multicolumn{1}{c}{\textbf{PASCAL Label}}   & \multicolumn{1}{c}{\textbf{Top Returned Credible Labels}} \\
\hline
 \multicolumn{1}{c}{aeroplane} & 'airplane, aeroplane, plane', 'jet, jet plane, jet-propelled plane', 'jetliner', 'warplane, military plane' \\
 \hline
 \multicolumn{1}{c}{bicycle} & 'bicycle, bike, wheel, cycle', 'safety bicycle, safety bike', 'push-bike', 'ordinary, ordinary bicycle' \\
 \hline
 \multicolumn{1}{c}{bird} & 'bird', 'passerine, passeriform bird', 'dickeybird, dickey-bird, dickybird, dicky-bird', 'parrot'\\
 \hline
 \multicolumn{1}{c}{boat} & 'boat', 'small boat', 'dinghy, dory, rowboat', 'sea boat', 'rowing boat', 'river boat', 'cockleshell'\\
  \hline
 \multicolumn{1}{c}{bottle} & 'bottle', 'pop bottle, soda bottle', 'water bottle', 'jar', 'smelling bottle', 'flask', 'jug', 'carafe'\\
   \hline
 \multicolumn{1}{c}{bus} & 'public transport', 'local', 'bus, autobus', 'express, limited', 'shuttle bus', 'trolleybus, trolley coach'\\
    \hline
 \multicolumn{1}{c}{car} & 'motor vehicle, automotive vehicle', 'car, auto', 'coupe', 'sports car, sport car', 'sedan, saloon' \\
     \hline
 \multicolumn{1}{c}{cat} & 'domestic cat, house cat', 'kitty, kitty-cat', 'tom, tomcat', 'mouser', 'Manx, Manx cat', 'tabby' \\
     \hline
 \multicolumn{1}{c}{chair} & 'chair', 'seat', 'armchair', 'straight chair, side chair', 'rocking chair, rocker', 'swivel chair'\\ 
      \hline
 \multicolumn{1}{c}{cow} & 'bull', 'cattle, cows', 'cow', 'bullock, steer', 'beef, beef cattle', 'cow, moo-cow', 'dairy cattle'\\ 
       \hline
 \multicolumn{1}{c}{dining table} & 'dining-room table', 'dining table, board', 'dinner table', 'table', 'dining-room furniture'\\        \hline
 \multicolumn{1}{c}{dog} & 'sporting dog, gun dog', 'terrier', 'retriever', 'hunting dog', 'Labrador retriever', 'water dog'\\
        \hline
 \multicolumn{1}{c}{horse} & 'horse, Equus caballus', 'equine, equid', 'gelding', 'mare, female horse', 'yearling', 'pony', 'dobbin'\\
 
        \hline
 \multicolumn{1}{c}{motorbike} & 'motorcycle, bike', 'wheeled vehicle', 'trail bike, dirt bike, scrambler', 'motor scooter, scooter'\\
   
           \hline
 \multicolumn{1}{c}{person} & 'person, individual', 'male, male person', 'face', 'oldster, old person', 'man', 'eccentric person'\\
   
           \hline
 \multicolumn{1}{c}{potted plant} & 'pot, flowerpot', 'planter', 'houseplant', 'bucket, pail', 'vase', 'crock, earthenware jar', 'watering pot'\\
 
            \hline
 \multicolumn{1}{c}{sheep} & 'sheep', 'domestic sheep, Ovis aries', 'ewe', 'ram, tup', 'black sheep', 'wild sheep', 'mountain sheep'\\

            \hline
 \multicolumn{1}{c}{sofa} & 'seat', 'sofa, couch, lounge', 'love seat, loveseat', 'chesterfield', 'settee', 'easy chair, lounge chair'\\
 
             \hline
 \multicolumn{1}{c}{train} & 'train, railroad train', 'passenger train', 'mail train', 'car train', 'freight train, rattler', 'commuter'\\
   
                \hline
 \multicolumn{1}{c}{tv/monitor} & 'monitor', 'LCD', 'television monitor, tv monitor', 'OLED', 'digital display, alphanumeric display'\\
   
\hline 
\end{tabular}
\label{tab:labels}
\end{table*}

\begin{table*}[htbp]
	\centering
	\caption{Evaluation on the quality of dataset construction.}
\label{tab:bbox_result}%
	\setlength\tabcolsep{3pt}
\resizebox{\linewidth}{!}{
	\begin{tabular}{|c|c|c|c|c|c|c|c|c|c|c|c|c|c|c|c|c|c|c|c|c|c|}
		\hline
		Method & \begin{sideways}aero\end{sideways} & \begin{sideways}bike\end{sideways} & \begin{sideways}bird\end{sideways} & \begin{sideways}boat\end{sideways} & \begin{sideways}bottle\end{sideways} & \begin{sideways}bus\end{sideways} & \begin{sideways}car\end{sideways} & \begin{sideways}cat\end{sideways} & \begin{sideways}chair\end{sideways} & \begin{sideways}cow\end{sideways} & \begin{sideways}table\end{sideways} & \begin{sideways}dog\end{sideways} & \begin{sideways}horse\end{sideways} &  \begin{sideways}mbike\end{sideways} & \begin{sideways}person\end{sideways} & \begin{sideways}plant\end{sideways} & \begin{sideways}sheep\end{sideways} &
	\begin{sideways}sofa\end{sideways} & \begin{sideways}train\end{sideways} &  
	\begin{sideways}tv\end{sideways} & Avg \\
	\hline
Retention Rate (deep) & 64.09 &  19.38 &  78.02 &  60.00 &  62.37 &  59.50 &  79.50 &  85.71 &  61.83 &  78.86 &  64.82 &  74.49 &  75.90 &  47.18 &  50.67 &  63.30 &  76.06 &  67.50 &  74.32 &  48.59 &  64.60  \\
Retention Rate (ebox) & 38.00 & 25.5 & 26.5 & 37.37 & 36.0 & 35.5 & 43.5 & 29.0 & 48.0 & 16.29 & 47.5 & 23.0 & 19.0 & 40.5 & 26.0 & 24.5 & 31.33 & 41.0 & 37.0 & 33.55 & 32.95  \\
\hline
FP Rate (deep) & 0.00 &  12.41 &  1.74 &  3.73 &  11.11 &  3.01 &  2.01 &  4.02 &  4.71 &  3.43 &  15.63 &  1.16 &  1.55 &  9.34 &  3.51 &  12.50 &  2.13 &  6.12 &  1.68 &  7.62 &  5.37 \\
FP Rate (ebox) & 4.5 & 11.5 & 19.79 & 11.16 & 9.64 & 2.5 & 3.51 & 9.54 & 6.63 & 9.55 & 13.5 & 14.19 & 13.0 & 12.0 & 14.06 & 27.71 & 15.64 & 11.16 & 6.03 & 18.86 & 11.73 \\
\hline
\end{tabular}
}%
\end{table*}%

\subsection{Dataset Construction}

\subsubsection{Experiment setup}

To evaluate our automatic dataset construction method, we use a pre-trained 11k-class classification model trained on the ImageNet dataset\footnote{We converted a pre-trained MXNet model to Pytorch \url{http://data.mxnet.io/models/imagenet-11k/}.}, and a word2vec model pretrained on Google News dataset (about 100 billion words) which contains 300-dimensional vectors for 3 million different words and phrases. 
To generate the noisy bounding boxes from an image, we have adopted two methods including:

\setlist{nolistsep}
\begin{itemize}[noitemsep]
    \item \textbf{deep:} We train an object detection model (RetinaNet) on the COCO dataset~\cite{lin2014microsoft} by excluding the 20 classes in Pascal VOC and the two new classes in iKitchen, i.e. ``slow cooker" and ``cocktail shaker". For each image, we run the detector to identify bounding boxes with classification confidence scores above a low threshold 0.2.
    \item \textbf{edge:} We run the EdgeBoxes~\cite{zitnick2014edge} algorithm on each image to retain up to 20 bounding boxes based on the predicted confidence scores.
\end{itemize}

\subsubsection{Pascal 20 Classes}

First, we would like to demonstrate that the set of ``credible labels" identified in the label set of the 11k classification model by the proposed Algorithm~\ref{alg:data} is reasonable. The produced credible labels are given in Table~\ref{tab:labels}. We can observe that our algorithm has managed to extract the semantically closest labels for the given class names. Some of the extracted class labels include rarely used words which may be ignored even by humans. These accurately identified ``credible labels" is critical for removing a large number of irrelevant bounding boxes from the noisy bounding box candidate set. 

% \textcolor{red}{(Dawei: number of boxes before and after)}

Second, we calculate the accuracy of the returned bounding boxes by comparing them with the ground-truth. Two evaluation metrics are used:

\setlist{nolistsep}
\begin{itemize}[noitemsep]
    \item \textbf{Retention Rate:} The retention rate is defined as the percentage of images that have been labeled correctly for all the ground-truth bounding boxes. Here, we say a box is correctly labeled if the returned bounding box has IoU (intersection over union: $\frac{area(Box_{gth})\cap area(Box_{prd}}{area(Box_{gth})\cup area(Box_{prd}}$) above 0.5 to the ground-truth bounding box.
    \item \textbf{FP Rate:} The false positive rate is defined as the percentage of bounding boxes that fall out of the ground-truth region. Concretely, we calculate the IoP value (intersection over prediction: $\frac{area(Box_{gth})\cap area(Box_{prd}}{area(Box_{prd}))}$) to decide if the predicted bounding box is within the scope of a ground-truth bounding box. If IoP<0.5, we regard this predicted bounding box as a false positive. A high false positive rate will have serious negative impact on the training accuracy.
\end{itemize}

We present the result of the bounding box extraction accuracy in Table~\ref{tab:bbox_result}. Compared with the noisy bounding boxes generated by EdgeBoxes, the boxes generated by the deep learning detection model are much more accurate in terms of both \textit{retention rate} and \textit{FP rate}. This can be explained by the fact that even though the deep learning model is not trained on the new classes, it has ``seen" tens of thousands of images, and thus could better identify the boundaries between different semantically meaningful objects with deep learned features. On the other hand, EdgeBoxes generates bounding box proposals that rely entirely on the edges extracted from the image, and thus is not as robust. On the noisy boxes generated by the deep learning model, our algorithm achieves a high retention rate of 64.6\% and low FP rate of 5.37\%. 

The only case with a low retention rate is the $bicycle$ class. After manually inspecting the generated bounding boxes, we observed that many bounding boxes were generated over the bicycle wheels instead of the whole bicycle. This happens because the classification model also classifies a bounding box with a bicycle wheel as ``bicycle". However, since those bounding boxes would fall within the boundary of the ground-truth bounding boxes, the FP rate for the bicycle class is still only 12.4\%.

% \textcolor{red}{(sample bounding boxes images)}

% Time vs number of images.

% We compare a deep model vs edge boxes methods.

% \textcolor{red}{(If time allowed: Train with generated boxes)}

\subsubsection{Incrementally Train iKitchen Dataset with Automatically Constructed Dataset}

\begin{table}[t]
\caption{iKitchen Accuracy on Automatically Constructed Dataset. ($Base_O$: average mAP for the base 8 classes before learning the new class; $Base_N$: average mAP for the base 8 classes after learning the new class; $N$: mAP for the new learned class; $Avg$: average mAP for 9 classes after learning the new class.)}
\centering
\footnotesize
\begin{tabular}{m{1.15cm}|m{1.2cm}|m{1.2cm}|m{1.2cm}|m{1.2cm}}
\hline
 & $Base_O$  & $Base_N$  & $New$ &  $Avg$ \\
\hline
8+SC & 80.1\% & 80.8\% & 85.4\% & 81.3\% \\
8+CS & 80.1\% & 79.2\% & 32.2\% & 74.0\% \\
\hline 
\end{tabular}
\label{tab:train_download_data}
\end{table}

For \textit{iKitchen} dataset, we download 100 images for the new class with Google image search\footnote{We only use images labeled strictly with the usage right ``free to use, share or modify, even commercially".} for incremental training. In particular, we use the keywords ``slow cooker" and ``cocktail shaker" for the 8+SC and 8+CS scenarios respectively. After running the dataset construction algorithm, 71 images remain for the ``slow cooker" class and 91 images remain for the ``cocktail shaker" class. We train the model with 10 exemplars per old class.

We show the result in Table~\ref{tab:train_download_data}. For both 8+SC and 8+CS scenarios, after learning the new class, the accuracy on the old classes almost remain unchanged. For the new learned class, we have achieved very good accuracy on the slow cooker class with mAP 85.4\%. However, for 8+CS scenario, the accuracy for the new ``cocktail shaker" class is relatively low with mAP 32.2\%.  This demonstrates that the quality of the automatically created training dataset is not consistent for different object classes. We will make further investigation on this issue in our future work.

\subsection{System Implementation and Efficiency}

To measure the latency of our incremental learning approach, we have implemented the end-to-end system and performed multiple experiments using two experimental setups. In \textit{Edge-Only} setup, all steps from dataset collection to incremental training of new classes is performed on the embedded Jetson TX2 platform. While this setup has advantages like privacy protection and reduced network activity, it cannot benefit from the powerful GPU(s) on the cloud side.  On the other hand, in \textit{Edge-Cloud} setup, the major system components including dataset preparation and incremental model training run on the cloud server which uses a single NVIDIA Tesla M40 GPU and then the final model is transferred to a Samsung Galaxy S9 Android phone for inference. Table \ref{tab:system_time} shows the latencies for every step of the incremental learning process in both setups. Even though Edge-Only setup has no model transfer latency, overall it is much slower than Edge-Cloud setup due to the significant difference in computation power during model training. Please refer to the video in the supplementary material for a demo of our Edge-Cloud implementation.

\begin{table}[t]
\caption{Training time of different input image size on NVIDIA Tesla M40 (seconds per epoch).}
\centering
\footnotesize
\begin{tabular}{m{2.6cm}|m{2.2cm}|m{2.2cm}}
\hline
Input Size&\textbf{10 Exemplar}&\textbf{30 Exemplar}\\
% \multicolumn{2}{c|}{\textbf{Mobile-Only}} & \multicolumn{2}{c}{\textbf{Mobile-Cloud}} \\
\hline
Small (512) & 8.3 & 14.1 \\
Large (1024) & 17.2 & 30.7 \\
\hline 
\end{tabular}
\label{tab:image_size}
\end{table}

% Measure latency for each component in dataset construction
% 10 epochs, 304size, 100 images, resnet18
\begin{table}[t]
\caption{System Running Time (s).}
\centering
\footnotesize
\begin{tabular}{m{2.4cm}|m{2.4cm}|m{2.4cm}}
\hline
&\textbf{Edge-Only}&\textbf{Edge-Cloud}\\
% \multicolumn{2}{c|}{\textbf{Mobile-Only}} & \multicolumn{2}{c}{\textbf{Mobile-Cloud}} \\
\hline
Download image & 16 & 10 \\
Build dataset & 44 & 21 \\
Train model & 233 & 83 \\
Download model & N/A & 5 \\
\hline
\textbf{Total}  & \textbf{293} & \textbf{119} \\
\hline 
\end{tabular}
\label{tab:system_time}
\end{table}

\begin{figure}[t]
\subfloat[10 Exemplars]{%
  \includegraphics[clip,width=0.9\columnwidth]{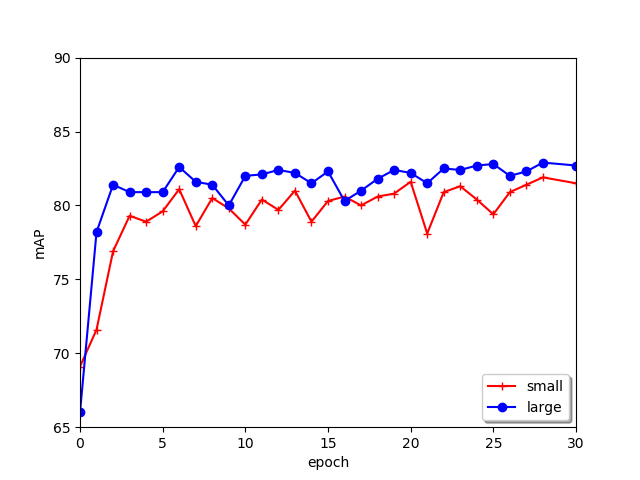}%
}
\vspace{-0.15in}
\subfloat[30 Exemplars]{%
  \includegraphics[clip,width=0.9\columnwidth]{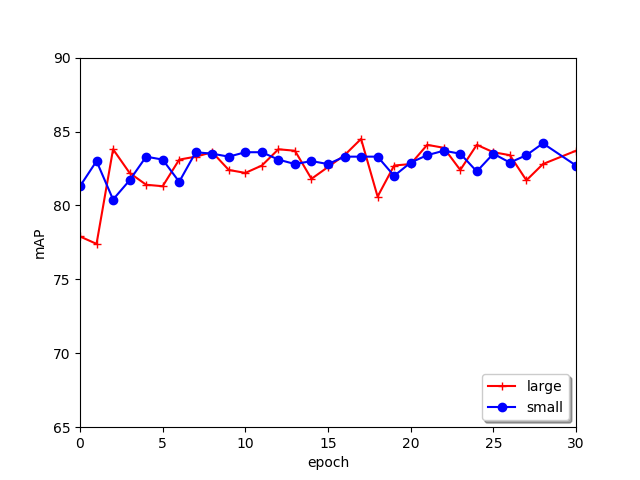}%
}
\caption{Accuracy of different input sizes for 8+slow cooker.}
\label{fig:map_exemplar}
\end{figure}

In these experiments, we used Resnet-18 as the base model, and trained the model for 10 epochs on automatically generated dataset from 100 images downloaded using Google image search plus 10 exemplar images per the base-8 class. In addition, the input image is resized that the longer side is 512 while maintaining the original aspect ratio. Note that we preferred a small image size since it has a major effect on training time. Table \ref{tab:image_size} shows that doubling the image dimensions more than doubles the training time while providing little improvement on the accuracy of the system (see Figure \ref{fig:map_exemplar}).

\subsection{Discussion}

% (\textcolor{red}{Shalini: Might be good to have an ablation section, where we discuss the degradation in results if we drop or simplify particular parts of our overall system (e.g., loss function terms, or semantic match of labels)  -- that would be a good way to motivate the overall system})
Even though the proposed incremental learning system for object detection is fast and practical enough to be deployed in real applications, many optimizations are possible to improve the system efficiency:
%In this paper, we tackled the general object detection in which there might be many objects of interest in an image with various sizes.  For such cases, the system can be optimized using the following design choices:

%\textcolor{red}{(Dawei: it is not clear how to use this to speedup our training.)}
%\textbf{Reduce the number of anchor boxes.} In Retinanet, our base object detector, thousands of anchor boxes are calculated using multiple layers in the network and for every spatial location on each layer's feature map. However, in many applications, the user is interested in a single salient object that is probably clearly visible and centered on the image. If we assume the object of interest is centered on the image, we can skip anchors that are closer to edges and corners of an image. In addition, if we assume the object should be relatively large on the image, we can skip anchor proposals from lower layers since they are used for detecting smaller objects on the image.

\textbf{Model optimizations} In this work, we used ResNet-50 and ResNet-18 as the backbone models. However, using a network that is primarily designed for time-critical edge applications such as MobileNet~\cite{howard2017mobilenets} can reduce the overall system time significantly by decreasing the model training time. 
% In addition, in single-shot object detectors, anchor boxes are calculated using multiple layers in the base feature extraction model.
% However, in many applications, the user is interested in a single salient object that is relatively large on the image. In such a case, we can skip anchor proposals from lower layers since they are particularly used for detecting smaller objects on the image.

\textbf{System enhancements.} Besides these model optimizations, when enough system resources are available system-level tricks such as, multi-gpu training, parallel image downloading and preprocessing, caching of downloaded images and pre-loading of deep models that are used in dataset generation can be employed to further reduce the overall system time.

% \textbf{Improve generated bounding box quality.} In the current implementation, we assume the training data for the new class are either readily available or not available at all which requires building a dataset on-the-fly. In the actual situation, it is very likely that we already have a small amount of perfectly labeled data. By incorporating the small set of data in our dataset construction pipeline, we could have constructed a training dataset with more reliable labels.

\textbf{Bounding box generation for random unseen objects.} In rare cases, the new objects a user is interested in might not be included in the large-scale classification model (e.g., 11k ImageNet). In such scenarios, we cannot use the proposed dataset preparation pipeline. Instead, we could possibly use unsupervised object co-localization method ~\cite{weideep} to locate objects belonging to the same category at the cost of less accurate bounding boxes locations.

\textbf{Scaling the model.} To maintain a good accuracy, exemplars from old classes must be included. However, this will also increase the model training time linearly. More intelligent methods of selecting the exemplars would help alleviate this problem.

\section{conclusion}
\label{sec:conclusion}

In this paper, we have presented \pname{}, an edge system for efficient incremental learning of deep neural networks for object detection. \pname{} has included a novel incremental learning algorithm which learns to detect a new object class with only training data from the new class while preventing the model from forgetting its knowledge on the old classes. Compared to the existing learning methods, this algorithm achieves 38x speed-up with negligible accuracy loss. In addition, in the absence of available training data for the new classes, \pname{} introduced a real-time dataset construction method to label web-crawled images with high-precision bounding boxes and this actually makes \pname{} a practical system ready for deployment. We have implemented \pname{} with both edge-cloud and edge-only architectures, and demonstrated that learning of a new object class can finish in less than 2 minutes on a single GPU with superior detection accuracy.

%In this paper, we present DeepMon, a system for enabling the
% low-latency execution of deep DNNs on a mobile GPU. DeepMon
% uses various optimisation techniques including the convolutional
% layer caching, decomposition, and matrix multiplication optimizations
% to achieve significant speedups (over 3-4x) compared
% to state-of-the-art current solutions. In particular, DeepMon allows
% DNN-based face/object detection models, such as VGG-VeryDeep16
% and YOLO, to process video frames at 1 to 2 frames per second.
% We implemented DeepMon in both OpenCL and Vulkan and
% tested it’s effectiveness using three different mobile GPUs (Adreno
% 420, Adreno 430, and Mali T 880) and against alternative solutions
% such as state-of-the-art research systems (DeepX) and plausible
% cloud-based solutions. Our results show that DeepMon significantly
% outperforms all local-computation-only based solutions
% with a marginal drop in accuracy and that DeepMon’s latency and
% accuracy combination can only be bettered by using very large
% (and expensive) cloud server instances with very good networking
% connectivity. Videos of DeepMon in action are available at
% http://is.gd/DeepMon. Also, DeepMon’s source code can be found
% at https://github.com/JC1DA/DeepMon.

\bibliographystyle{ACM-Reference-Format}
\bibliography{main}


\begin{thebibliography}{00}

%%% the Web address.

\ifx \showCODEN    \undefined \def \showCODEN     #1{\unskip}     \fi
\ifx \showDOI      \undefined \def \showDOI       #1{#1}\fi
\ifx \showISBNx    \undefined \def \showISBNx     #1{\unskip}     \fi
\ifx \showISBNxiii \undefined \def \showISBNxiii  #1{\unskip}     \fi
\ifx \showISSN     \undefined \def \showISSN      #1{\unskip}     \fi
\ifx \showLCCN     \undefined \def \showLCCN      #1{\unskip}     \fi
\ifx \shownote     \undefined \def \shownote      #1{#1}          \fi
\ifx \showarticletitle \undefined \def \showarticletitle #1{#1}   \fi
\ifx \showURL      \undefined \def \showURL       {\relax}        \fi
% invisible to TeX
\providecommand\bibfield[2]{#2}
\providecommand\bibinfo[2]{#2}
\providecommand\natexlab[1]{#1}
\providecommand\showeprint[2][]{arXiv:#2}

\bibitem[\protect\citeauthoryear{Abadi, Barham, Chen, Chen, Davis, Dean, Devin,
  Ghemawat, Irving, Isard, Kudlur, Levenberg, Monga, Moore, Murray, Steiner,
  Tucker, Vasudevan, Warden, Wicke, Yu, and Zheng}{Abadi et~al\mbox{.}}{2016}]%
        {tensorFlow}
\bibfield{author}{\bibinfo{person}{Mart{\'\i}n Abadi}, \bibinfo{person}{Paul
  Barham}, \bibinfo{person}{Jianmin Chen}, \bibinfo{person}{Zhifeng Chen},
  \bibinfo{person}{Andy Davis}, \bibinfo{person}{Jeffrey Dean},
  \bibinfo{person}{Matthieu Devin}, \bibinfo{person}{Sanjay Ghemawat},
  \bibinfo{person}{Geoffrey Irving}, \bibinfo{person}{Michael Isard},
  \bibinfo{person}{Manjunath Kudlur}, \bibinfo{person}{Josh Levenberg},
  \bibinfo{person}{Rajat Monga}, \bibinfo{person}{Sherry Moore},
  \bibinfo{person}{Derek~G. Murray}, \bibinfo{person}{Benoit Steiner},
  \bibinfo{person}{Paul Tucker}, \bibinfo{person}{Vijay Vasudevan},
  \bibinfo{person}{Pete Warden}, \bibinfo{person}{Martin Wicke},
  \bibinfo{person}{Yuan Yu}, {and} \bibinfo{person}{Xiaoqiang Zheng}.}
  \bibinfo{year}{2016}\natexlab{}.
\newblock \showarticletitle{TensorFlow: A System for Large-Scale Machine
  Learning}. In \bibinfo{booktitle}{{\em 12th USENIX Symposium on Operating
  Systems Design and Implementation (OSDI'16)}}. \bibinfo{address}{Savannah,
  GA}.
\newblock


\bibitem[\protect\citeauthoryear{Aljundi, Babiloni, Elhoseiny, Rohrbach, and
  Tuytelaars}{Aljundi et~al\mbox{.}}{2017}]%
        {aljundi2017memory}
\bibfield{author}{\bibinfo{person}{Rahaf Aljundi}, \bibinfo{person}{Francesca
  Babiloni}, \bibinfo{person}{Mohamed Elhoseiny}, \bibinfo{person}{Marcus
  Rohrbach}, {and} \bibinfo{person}{Tinne Tuytelaars}.}
  \bibinfo{year}{2017}\natexlab{}.
\newblock \showarticletitle{Memory Aware Synapses: Learning what (not) to
  forget}.
\newblock \bibinfo{journal}{{\em arXiv preprint arXiv:1711.09601\/}}
  (\bibinfo{year}{2017}).
\newblock


\bibitem[\protect\citeauthoryear{Chen, Choi, Yu, Han, and Chandraker}{Chen
  et~al\mbox{.}}{2017}]%
        {chen2017learning}
\bibfield{author}{\bibinfo{person}{Guobin Chen}, \bibinfo{person}{Wongun Choi},
  \bibinfo{person}{Xiang Yu}, \bibinfo{person}{Tony Han}, {and}
  \bibinfo{person}{Manmohan Chandraker}.} \bibinfo{year}{2017}\natexlab{}.
\newblock \showarticletitle{Learning efficient object detection models with
  knowledge distillation}. In \bibinfo{booktitle}{{\em Advances in Neural
  Information Processing Systems}}. \bibinfo{pages}{742--751}.
\newblock


\bibitem[\protect\citeauthoryear{Chong, Blei, and Li}{Chong
  et~al\mbox{.}}{2009}]%
        {chong2009simultaneous}
\bibfield{author}{\bibinfo{person}{Wang Chong}, \bibinfo{person}{David Blei},
  {and} \bibinfo{person}{Fei-Fei Li}.} \bibinfo{year}{2009}\natexlab{}.
\newblock \showarticletitle{Simultaneous image classification and annotation}.
  In \bibinfo{booktitle}{{\em Computer Vision and Pattern Recognition, 2009.
  CVPR 2009. IEEE Conference on}}. IEEE, \bibinfo{pages}{1903--1910}.
\newblock


\bibitem[\protect\citeauthoryear{Deng, Dong, Socher, Li, Li, and Fei-Fei}{Deng
  et~al\mbox{.}}{2009}]%
        {imagenet_cvpr09}
\bibfield{author}{\bibinfo{person}{J. Deng}, \bibinfo{person}{W. Dong},
  \bibinfo{person}{R. Socher}, \bibinfo{person}{L.-J. Li}, \bibinfo{person}{K.
  Li}, {and} \bibinfo{person}{L. Fei-Fei}.} \bibinfo{year}{2009}\natexlab{}.
\newblock \showarticletitle{{ImageNet: A Large-Scale Hierarchical Image
  Database}}. In \bibinfo{booktitle}{{\em CVPR09}}.
\newblock


\bibitem[\protect\citeauthoryear{Everingham, Van~Gool, Williams, Winn, and
  Zisserman}{Everingham et~al\mbox{.}}{2010}]%
        {everingham2010pascal}
\bibfield{author}{\bibinfo{person}{Mark Everingham}, \bibinfo{person}{Luc
  Van~Gool}, \bibinfo{person}{Christopher~KI Williams}, \bibinfo{person}{John
  Winn}, {and} \bibinfo{person}{Andrew Zisserman}.}
  \bibinfo{year}{2010}\natexlab{}.
\newblock \showarticletitle{The pascal visual object classes (voc) challenge}.
\newblock \bibinfo{journal}{{\em International journal of computer vision\/}}
  \bibinfo{volume}{88}, \bibinfo{number}{2} (\bibinfo{year}{2010}),
  \bibinfo{pages}{303--338}.
\newblock


\bibitem[\protect\citeauthoryear{Everingham, Van~Gool, Williams, Winn, and
  Zisserman}{Everingham et~al\mbox{.}}{[n. d.]}]%
        {pascal-voc-2007}
\bibfield{author}{\bibinfo{person}{M. Everingham}, \bibinfo{person}{L.
  Van~Gool}, \bibinfo{person}{C.~K.~I. Williams}, \bibinfo{person}{J. Winn},
  {and} \bibinfo{person}{A. Zisserman}.} \bibinfo{year}{[n. d.]}\natexlab{}.
\newblock \bibinfo{title}{The {PASCAL} {V}isual {O}bject {C}lasses {C}hallenge
  2007 {(VOC2007)} {R}esults}.
\newblock
  \bibinfo{howpublished}{http://www.pascal-network.org/challenges/VOC/voc2007/workshop/index.html}.
    (\bibinfo{year}{[n. d.]}).
\newblock


\bibitem[\protect\citeauthoryear{Girshick}{Girshick}{2015}]%
        {girshick2015fast}
\bibfield{author}{\bibinfo{person}{Ross Girshick}.}
  \bibinfo{year}{2015}\natexlab{}.
\newblock \showarticletitle{Fast r-cnn}. In \bibinfo{booktitle}{{\em
  Proceedings of the IEEE international conference on computer vision}}.
  \bibinfo{pages}{1440--1448}.
\newblock


\bibitem[\protect\citeauthoryear{Han, Mao, and Dally}{Han
  et~al\mbox{.}}{2015}]%
        {han_compression}
\bibfield{author}{\bibinfo{person}{Song Han}, \bibinfo{person}{Huizi Mao},
  {and} \bibinfo{person}{William~J. Dally}.} \bibinfo{year}{2015}\natexlab{}.
\newblock \showarticletitle{{Deep Compression: Compressing Deep Neural Network
  with Pruning, Trained Quantization and Huffman Coding}}.
\newblock \bibinfo{journal}{{\em CoRR\/}}  \bibinfo{volume}{abs/1510.00149}
  (\bibinfo{year}{2015}).
\newblock


\bibitem[\protect\citeauthoryear{Han, Shen, Philipose, Agarwal, Wolman, and
  Krishnamurthy}{Han et~al\mbox{.}}{2016}]%
        {han:mobisys2016}
\bibfield{author}{\bibinfo{person}{Seungyeop Han}, \bibinfo{person}{Haichen
  Shen}, \bibinfo{person}{Matthai Philipose}, \bibinfo{person}{Sharad Agarwal},
  \bibinfo{person}{Alec Wolman}, {and} \bibinfo{person}{Arvind Krishnamurthy}.}
  \bibinfo{year}{2016}\natexlab{}.
\newblock \showarticletitle{{MCDNN: An Approximation-Based Execution Framework
  for Deep Stream Processing Under Resource Constrains}}. In
  \bibinfo{booktitle}{{\em Proc. MobiSys'16}}. \bibinfo{address}{Singapore}.
\newblock


\bibitem[\protect\citeauthoryear{He, Zhang, Ren, and Sun}{He
  et~al\mbox{.}}{2015}]%
        {resnet}
\bibfield{author}{\bibinfo{person}{Kaiming He}, \bibinfo{person}{Xiangyu
  Zhang}, \bibinfo{person}{Shaoqing Ren}, {and} \bibinfo{person}{Jian Sun}.}
  \bibinfo{year}{2015}\natexlab{}.
\newblock \showarticletitle{{Deep Residual Learning for Image Recognition}}.
\newblock \bibinfo{journal}{{\em CoRR\/}}  \bibinfo{volume}{abs/1512.03385}
  (\bibinfo{year}{2015}).
\newblock


\bibitem[\protect\citeauthoryear{Hinton, Vinyals, and Dean}{Hinton
  et~al\mbox{.}}{2015}]%
        {hinton2015distilling}
\bibfield{author}{\bibinfo{person}{Geoffrey Hinton}, \bibinfo{person}{Oriol
  Vinyals}, {and} \bibinfo{person}{Jeff Dean}.}
  \bibinfo{year}{2015}\natexlab{}.
\newblock \showarticletitle{Distilling the knowledge in a neural network}.
\newblock \bibinfo{journal}{{\em arXiv preprint arXiv:1503.02531\/}}
  (\bibinfo{year}{2015}).
\newblock


\bibitem[\protect\citeauthoryear{Howard, Zhu, Chen, Kalenichenko, Wang, Weyand,
  Andreetto, and Adam}{Howard et~al\mbox{.}}{2017}]%
        {howard2017mobilenets}
\bibfield{author}{\bibinfo{person}{Andrew~G Howard}, \bibinfo{person}{Menglong
  Zhu}, \bibinfo{person}{Bo Chen}, \bibinfo{person}{Dmitry Kalenichenko},
  \bibinfo{person}{Weijun Wang}, \bibinfo{person}{Tobias Weyand},
  \bibinfo{person}{Marco Andreetto}, {and} \bibinfo{person}{Hartwig Adam}.}
  \bibinfo{year}{2017}\natexlab{}.
\newblock \showarticletitle{Mobilenets: Efficient convolutional neural networks
  for mobile vision applications}.
\newblock \bibinfo{journal}{{\em arXiv preprint arXiv:1704.04861\/}}
  (\bibinfo{year}{2017}).
\newblock


\bibitem[\protect\citeauthoryear{Huang, Liu, van~der Maaten, and
  Weinberger}{Huang et~al\mbox{.}}{2017}]%
        {huang2017densely}
\bibfield{author}{\bibinfo{person}{Gao Huang}, \bibinfo{person}{Zhuang Liu},
  \bibinfo{person}{Laurens van~der Maaten}, {and} \bibinfo{person}{Kilian~Q
  Weinberger}.} \bibinfo{year}{2017}\natexlab{}.
\newblock \showarticletitle{Densely Connected Convolutional Networks}. In
  \bibinfo{booktitle}{{\em 2017 IEEE Conference on Computer Vision and Pattern
  Recognition (CVPR)}}. IEEE, \bibinfo{pages}{2261--2269}.
\newblock


\bibitem[\protect\citeauthoryear{Huynh, Lee, and Balan}{Huynh
  et~al\mbox{.}}{2017}]%
        {huynh2017deepmon}
\bibfield{author}{\bibinfo{person}{Loc~N Huynh}, \bibinfo{person}{Youngki Lee},
  {and} \bibinfo{person}{Rajesh~Krishna Balan}.}
  \bibinfo{year}{2017}\natexlab{}.
\newblock \showarticletitle{Deepmon: Mobile gpu-based deep learning framework
  for continuous vision applications}. In \bibinfo{booktitle}{{\em Proceedings
  of the 15th Annual International Conference on Mobile Systems, Applications,
  and Services}}. ACM, \bibinfo{pages}{82--95}.
\newblock


\bibitem[\protect\citeauthoryear{Iandola, Moskewicz, Ashraf, Han, Dally, and
  Keutzer}{Iandola et~al\mbox{.}}{2016}]%
        {squeezenet}
\bibfield{author}{\bibinfo{person}{Forrest~N. Iandola},
  \bibinfo{person}{Matthew~W. Moskewicz}, \bibinfo{person}{Khalid Ashraf},
  \bibinfo{person}{Song Han}, \bibinfo{person}{William~J. Dally}, {and}
  \bibinfo{person}{Kurt Keutzer}.} \bibinfo{year}{2016}\natexlab{}.
\newblock \showarticletitle{{SqueezeNet: AlexNet-level Accuracy with 50x Fewer
  Parameters and 1MB Model Size}}.
\newblock \bibinfo{journal}{{\em CoRR\/}}  \bibinfo{volume}{abs/1602.07360}
  (\bibinfo{year}{2016}).
\newblock


\bibitem[\protect\citeauthoryear{Jeon, Lavrenko, and Manmatha}{Jeon
  et~al\mbox{.}}{2003}]%
        {jeon2003automatic}
\bibfield{author}{\bibinfo{person}{Jiwoon Jeon}, \bibinfo{person}{Victor
  Lavrenko}, {and} \bibinfo{person}{Raghavan Manmatha}.}
  \bibinfo{year}{2003}\natexlab{}.
\newblock \showarticletitle{Automatic image annotation and retrieval using
  cross-media relevance models}. In \bibinfo{booktitle}{{\em Proceedings of the
  26th annual international ACM SIGIR conference on Research and development in
  informaion retrieval}}. ACM, \bibinfo{pages}{119--126}.
\newblock


\bibitem[\protect\citeauthoryear{Kingma and Ba}{Kingma and Ba}{2015}]%
        {kingma2015adam}
\bibfield{author}{\bibinfo{person}{Diederik~P. Kingma} {and}
  \bibinfo{person}{Jimmy Ba}.} \bibinfo{year}{2015}\natexlab{}.
\newblock \showarticletitle{Adam: A method for stochastic optimization}. In
  \bibinfo{booktitle}{{\em International Conference on Learning Representations
  (ICLR)}}.
\newblock


\bibitem[\protect\citeauthoryear{Kirkpatrick, Pascanu, Rabinowitz, Veness,
  Desjardins, Rusu, Milan, Quan, Ramalho, Grabska-Barwinska,
  et~al\mbox{.}}{Kirkpatrick et~al\mbox{.}}{2017}]%
        {kirkpatrick2017overcoming}
\bibfield{author}{\bibinfo{person}{James Kirkpatrick}, \bibinfo{person}{Razvan
  Pascanu}, \bibinfo{person}{Neil Rabinowitz}, \bibinfo{person}{Joel Veness},
  \bibinfo{person}{Guillaume Desjardins}, \bibinfo{person}{Andrei~A Rusu},
  \bibinfo{person}{Kieran Milan}, \bibinfo{person}{John Quan},
  \bibinfo{person}{Tiago Ramalho}, \bibinfo{person}{Agnieszka
  Grabska-Barwinska}, {et~al\mbox{.}}} \bibinfo{year}{2017}\natexlab{}.
\newblock \showarticletitle{Overcoming catastrophic forgetting in neural
  networks}.
\newblock \bibinfo{journal}{{\em Proceedings of the national academy of
  sciences\/}} (\bibinfo{year}{2017}), \bibinfo{pages}{201611835}.
\newblock


\bibitem[\protect\citeauthoryear{Lane, Bhattacharya, Georgiev, Forlivesi, Jiao,
  Qendro, and Kawsar}{Lane et~al\mbox{.}}{2016}]%
        {lane:ipsn2016}
\bibfield{author}{\bibinfo{person}{Nicholas~D. Lane}, \bibinfo{person}{Sourav
  Bhattacharya}, \bibinfo{person}{Petko Georgiev}, \bibinfo{person}{Claudio
  Forlivesi}, \bibinfo{person}{Lei Jiao}, \bibinfo{person}{Lorena Qendro},
  {and} \bibinfo{person}{Fahim Kawsar}.} \bibinfo{year}{2016}\natexlab{}.
\newblock \showarticletitle{{DeepX: A Software Accelerator for Low-Power Deep
  Learning Inference on Mobile Devices}}. In \bibinfo{booktitle}{{\em Proc.
  IPSN'16}}. \bibinfo{address}{Vienna, Austria}.
\newblock


\bibitem[\protect\citeauthoryear{Lane and Georgiev}{Lane and Georgiev}{2015}]%
        {lane:hotmobile2015}
\bibfield{author}{\bibinfo{person}{Nicholas~D. Lane} {and}
  \bibinfo{person}{Petko Georgiev}.} \bibinfo{year}{2015}\natexlab{}.
\newblock \showarticletitle{{Can Deep Learning Revolutionize Mobile Sensing?}}.
  In \bibinfo{booktitle}{{\em Proc. HotMobile'15}}. \bibinfo{address}{Santa Fe,
  NM}.
\newblock


\bibitem[\protect\citeauthoryear{Li and Hoiem}{Li and Hoiem}{2017}]%
        {li2017learning}
\bibfield{author}{\bibinfo{person}{Zhizhong Li} {and} \bibinfo{person}{Derek
  Hoiem}.} \bibinfo{year}{2017}\natexlab{}.
\newblock \showarticletitle{Learning without forgetting}.
\newblock \bibinfo{journal}{{\em IEEE Transactions on Pattern Analysis and
  Machine Intelligence\/}} (\bibinfo{year}{2017}).
\newblock


\bibitem[\protect\citeauthoryear{Lin, Doll{\'a}r, Girshick, He, Hariharan, and
  Belongie}{Lin et~al\mbox{.}}{2017}]%
        {lin2017feature}
\bibfield{author}{\bibinfo{person}{Tsung-Yi Lin}, \bibinfo{person}{Piotr
  Doll{\'a}r}, \bibinfo{person}{Ross Girshick}, \bibinfo{person}{Kaiming He},
  \bibinfo{person}{Bharath Hariharan}, {and} \bibinfo{person}{Serge Belongie}.}
  \bibinfo{year}{2017}\natexlab{}.
\newblock \showarticletitle{Feature Pyramid Networks for Object Detection}. In
  \bibinfo{booktitle}{{\em Computer Vision and Pattern Recognition (CVPR), 2017
  IEEE Conference on}}. IEEE, \bibinfo{pages}{936--944}.
\newblock


\bibitem[\protect\citeauthoryear{Lin, Goyal, Girshick, He, and Doll{\'a}r}{Lin
  et~al\mbox{.}}{2018}]%
        {lin2018focal}
\bibfield{author}{\bibinfo{person}{Tsung-Yi Lin}, \bibinfo{person}{Priyal
  Goyal}, \bibinfo{person}{Ross Girshick}, \bibinfo{person}{Kaiming He}, {and}
  \bibinfo{person}{Piotr Doll{\'a}r}.} \bibinfo{year}{2018}\natexlab{}.
\newblock \showarticletitle{Focal loss for dense object detection}.
\newblock \bibinfo{journal}{{\em IEEE transactions on pattern analysis and
  machine intelligence\/}} (\bibinfo{year}{2018}).
\newblock


\bibitem[\protect\citeauthoryear{Lin, Maire, Belongie, Hays, Perona, Ramanan,
  Doll{\'a}r, and Zitnick}{Lin et~al\mbox{.}}{2014}]%
        {lin2014microsoft}
\bibfield{author}{\bibinfo{person}{Tsung-Yi Lin}, \bibinfo{person}{Michael
  Maire}, \bibinfo{person}{Serge Belongie}, \bibinfo{person}{James Hays},
  \bibinfo{person}{Pietro Perona}, \bibinfo{person}{Deva Ramanan},
  \bibinfo{person}{Piotr Doll{\'a}r}, {and} \bibinfo{person}{C~Lawrence
  Zitnick}.} \bibinfo{year}{2014}\natexlab{}.
\newblock \showarticletitle{Microsoft coco: Common objects in context}. In
  \bibinfo{booktitle}{{\em European conference on computer vision}}. Springer,
  \bibinfo{pages}{740--755}.
\newblock


\bibitem[\protect\citeauthoryear{Liu, Anguelov, Erhan, Szegedy, Reed, Fu, and
  Berg}{Liu et~al\mbox{.}}{2016}]%
        {liu2016ssd}
\bibfield{author}{\bibinfo{person}{Wei Liu}, \bibinfo{person}{Dragomir
  Anguelov}, \bibinfo{person}{Dumitru Erhan}, \bibinfo{person}{Christian
  Szegedy}, \bibinfo{person}{Scott Reed}, \bibinfo{person}{Cheng-Yang Fu},
  {and} \bibinfo{person}{Alexander~C Berg}.} \bibinfo{year}{2016}\natexlab{}.
\newblock \showarticletitle{Ssd: Single shot multibox detector}. In
  \bibinfo{booktitle}{{\em European conference on computer vision}}. Springer,
  \bibinfo{pages}{21--37}.
\newblock


\bibitem[\protect\citeauthoryear{Mikolov, Sutskever, Chen, Corrado, and
  Dean}{Mikolov et~al\mbox{.}}{2013}]%
        {mikolov2013distributed}
\bibfield{author}{\bibinfo{person}{Tomas Mikolov}, \bibinfo{person}{Ilya
  Sutskever}, \bibinfo{person}{Kai Chen}, \bibinfo{person}{Greg~S Corrado},
  {and} \bibinfo{person}{Jeff Dean}.} \bibinfo{year}{2013}\natexlab{}.
\newblock \showarticletitle{Distributed representations of words and phrases
  and their compositionality}. In \bibinfo{booktitle}{{\em Advances in neural
  information processing systems}}. \bibinfo{pages}{3111--3119}.
\newblock


\bibitem[\protect\citeauthoryear{Ng}{Ng}{2017}]%
        {ng2017machine}
\bibfield{author}{\bibinfo{person}{Andrew Ng}.}
  \bibinfo{year}{2017}\natexlab{}.
\newblock \bibinfo{title}{Machine Learning Yearning}.
\newblock   (\bibinfo{year}{2017}).
\newblock


\bibitem[\protect\citeauthoryear{Paszke, Gross, Chintala, and Chanan}{Paszke
  et~al\mbox{.}}{2017}]%
        {paszke2017pytorch}
\bibfield{author}{\bibinfo{person}{Adam Paszke}, \bibinfo{person}{Sam Gross},
  \bibinfo{person}{Soumith Chintala}, {and} \bibinfo{person}{Gregory Chanan}.}
  \bibinfo{year}{2017}\natexlab{}.
\newblock \bibinfo{title}{Pytorch: Tensors and dynamic neural networks in
  python with strong gpu acceleration}.
\newblock   (\bibinfo{year}{2017}).
\newblock


\bibitem[\protect\citeauthoryear{{Porting Caffe to Android Platform}}{{Porting
  Caffe to Android Platform}}{[n. d.]}]%
        {CaffeAndroid}
\bibfield{author}{\bibinfo{person}{{Porting Caffe to Android Platform}}.}
  \bibinfo{year}{[n. d.]}\natexlab{}.
\newblock
  \bibinfo{howpublished}{\\\url{https://github.com/sh1r0/caffe-android-lib}}.
  (\bibinfo{year}{[n. d.]}).
\newblock


\bibitem[\protect\citeauthoryear{Rebuffi, Kolesnikov, Sperl, and
  Lampert}{Rebuffi et~al\mbox{.}}{2017}]%
        {rebuffi2017icarl}
\bibfield{author}{\bibinfo{person}{Sylvestre-Alvise Rebuffi},
  \bibinfo{person}{Alexander Kolesnikov}, \bibinfo{person}{Georg Sperl}, {and}
  \bibinfo{person}{Christoph~H Lampert}.} \bibinfo{year}{2017}\natexlab{}.
\newblock \showarticletitle{icarl: Incremental classifier and representation
  learning}. In \bibinfo{booktitle}{{\em Proceedings of the IEEE Conference on
  Computer Vision and Pattern Recognition}}.
\newblock


\bibitem[\protect\citeauthoryear{Redmon, Divvala, Girshick, and Farhadi}{Redmon
  et~al\mbox{.}}{2016}]%
        {redmon2016you}
\bibfield{author}{\bibinfo{person}{Joseph Redmon}, \bibinfo{person}{Santosh
  Divvala}, \bibinfo{person}{Ross Girshick}, {and} \bibinfo{person}{Ali
  Farhadi}.} \bibinfo{year}{2016}\natexlab{}.
\newblock \showarticletitle{You only look once: Unified, real-time object
  detection}. In \bibinfo{booktitle}{{\em Proceedings of the IEEE conference on
  computer vision and pattern recognition}}. \bibinfo{pages}{779--788}.
\newblock


\bibitem[\protect\citeauthoryear{Ren, He, Girshick, and Sun}{Ren
  et~al\mbox{.}}{2015}]%
        {fasterrcnn}
\bibfield{author}{\bibinfo{person}{Shaoqing Ren}, \bibinfo{person}{Kaiming He},
  \bibinfo{person}{Ross Girshick}, {and} \bibinfo{person}{Jian Sun}.}
  \bibinfo{year}{2015}\natexlab{}.
\newblock \showarticletitle{Faster R-CNN: Towards Real-Time Object Detection
  with Region Proposal Networks}. In \bibinfo{booktitle}{{\em Proc. NIPS'15}}.
  \bibinfo{address}{Montreal, Canada}.
\newblock


\bibitem[\protect\citeauthoryear{Shmelkov, Schmid, and Alahari}{Shmelkov
  et~al\mbox{.}}{2017}]%
        {shmelkov2017incremental}
\bibfield{author}{\bibinfo{person}{Konstantin Shmelkov},
  \bibinfo{person}{Cordelia Schmid}, {and} \bibinfo{person}{Karteek Alahari}.}
  \bibinfo{year}{2017}\natexlab{}.
\newblock \showarticletitle{Incremental learning of object detectors without
  catastrophic forgetting}.
\newblock \bibinfo{journal}{{\em arXiv preprint arXiv:1708.06977\/}}
  (\bibinfo{year}{2017}).
\newblock


\bibitem[\protect\citeauthoryear{{The SDK for Jetpac's iOS Deep Belief Image
  Recognition Framework}}{{The SDK for Jetpac's iOS Deep Belief Image
  Recognition Framework}}{[n. d.]}]%
        {DeepBeliefSDK}
\bibfield{author}{\bibinfo{person}{{The SDK for Jetpac's iOS Deep Belief Image
  Recognition Framework}}.} \bibinfo{year}{[n. d.]}\natexlab{}.
\newblock
  \bibinfo{howpublished}{\\\url{https://github.com/jetpacapp/DeepBeliefSDK}}.
  (\bibinfo{year}{[n. d.]}).
\newblock


\bibitem[\protect\citeauthoryear{Uijlings, van~de Sande, Gevers, and
  Smeulders}{Uijlings et~al\mbox{.}}{2013}]%
        {selectiveSearch}
\bibfield{author}{\bibinfo{person}{J.~R.~R. Uijlings},
  \bibinfo{person}{K.~E.~A. van~de Sande}, \bibinfo{person}{T. Gevers}, {and}
  \bibinfo{person}{A.~W.~M. Smeulders}.} \bibinfo{year}{2013}\natexlab{}.
\newblock \showarticletitle{Selective Search for Object Recognition}.
\newblock \bibinfo{journal}{{\em International Journal of Computer Vision\/}}
  \bibinfo{volume}{104}, \bibinfo{number}{2} (\bibinfo{year}{2013}),
  \bibinfo{pages}{154--171}.
\newblock
\showURL{%
\url{https://ivi.fnwi.uva.nl/isis/publications/2013/UijlingsIJCV2013}}


\bibitem[\protect\citeauthoryear{Wang, Gong, Yu, Li, Xie, and Zhou}{Wang
  et~al\mbox{.}}{2017}]%
        {wang2017dlau}
\bibfield{author}{\bibinfo{person}{Chao Wang}, \bibinfo{person}{Lei Gong},
  \bibinfo{person}{Qi Yu}, \bibinfo{person}{Xi Li}, \bibinfo{person}{Yuan Xie},
  {and} \bibinfo{person}{Xuehai Zhou}.} \bibinfo{year}{2017}\natexlab{}.
\newblock \showarticletitle{DLAU: A scalable deep learning accelerator unit on
  FPGA}.
\newblock \bibinfo{journal}{{\em IEEE Transactions on Computer-Aided Design of
  Integrated Circuits and Systems\/}} \bibinfo{volume}{36}, \bibinfo{number}{3}
  (\bibinfo{year}{2017}), \bibinfo{pages}{513--517}.
\newblock


\bibitem[\protect\citeauthoryear{Wei, Zhang, Li, Xie, Wu, Shen, and Zhou}{Wei
  et~al\mbox{.}}{[n. d.]}]%
        {weideep}
\bibfield{author}{\bibinfo{person}{Xiu-Shen Wei}, \bibinfo{person}{Chen-Lin
  Zhang}, \bibinfo{person}{Yao Li}, \bibinfo{person}{Chen-Wei Xie},
  \bibinfo{person}{Jianxin Wu}, \bibinfo{person}{Chunhua Shen}, {and}
  \bibinfo{person}{Zhi-Hua Zhou}.} \bibinfo{year}{[n. d.]}\natexlab{}.
\newblock \showarticletitle{Deep Descriptor Transforming for Image
  Co-Localization}.
\newblock  (\bibinfo{year}{[n. d.]}).
\newblock


\bibitem[\protect\citeauthoryear{Zitnick and Doll{\'a}r}{Zitnick and
  Doll{\'a}r}{2014}]%
        {zitnick2014edge}
\bibfield{author}{\bibinfo{person}{C~Lawrence Zitnick} {and}
  \bibinfo{person}{Piotr Doll{\'a}r}.} \bibinfo{year}{2014}\natexlab{}.
\newblock \showarticletitle{Edge boxes: Locating object proposals from edges}.
  In \bibinfo{booktitle}{{\em European conference on computer vision}}.
  Springer, \bibinfo{pages}{391--405}.
\newblock


\end{thebibliography}

\end{document}